\documentclass{article}
\pdfoutput=1

\usepackage{microtype}
\usepackage{graphicx}
\usepackage{booktabs}
\usepackage[pdftitle={Grounded autonomous research: a fault-tolerant LLM pipeline from corpus to manuscript in frontier computational physics},pdfauthor={Haonan Huang},pdfsubject={},pdfkeywords={},pdfcreator={},pdfproducer={}]{hyperref}

\usepackage[accepted]{icml2026}

\makeatletter
\renewcommand{\Notice@String}{Accepted at ICML 2026 AI for Science Workshop.}
\makeatother

\usepackage{amsmath}
\usepackage{amssymb}
\usepackage{multirow}
\usepackage{array}
\usepackage{enumitem}
\usepackage{listings}
\usepackage{pdfpages}
\usepackage{url}
\usepackage{xcolor}

\graphicspath{{figures/}}

\icmltitlerunning{Grounded autonomous research}

\begin{document}

\twocolumn[
\icmltitle{Grounded autonomous research: a fault-tolerant LLM pipeline from corpus to manuscript in frontier computational physics}

\icmlsetsymbol{equal}{*}

\begin{icmlauthorlist}
\icmlauthor{Haonan Huang}{pu}
\end{icmlauthorlist}

\icmlaffiliation{pu}{Department of Physics, Princeton University, Princeton, NJ, USA}

\icmlcorrespondingauthor{Haonan Huang}{hnhuang@princeton.edu}

\icmlkeywords{Autonomous research, LLM agents, AI for science, Computational physics, Altermagnetism}

\vskip 0.3in
]

\printAffiliationsAndNotice{}

\begin{abstract}
Autonomous-research agents have demonstrated end-to-end LLM automation in machine-learning sandboxes where execution provides calibration. Frontier physical science differs categorically: physical reasoning underlies every methodology choice, toolchains are often underdocumented, and calibration must come from external literature anchors --- which unscaffolded agents cite but do not confront, hallucinating plausible, unverifiable results from internal priors. We present a pipeline that runs end-to-end from a corpus of 11{,}083 recent condensed-matter physics arXiv papers to a publication-grade manuscript with three substantive physics findings (here on altermagnetic piezomagnetism): the agent autonomously conceives a research direction by mapping the corpus, calibrates methodology by reproducing published references, conducts novel first-principles computations, and writes the manuscript --- grounded in literature throughout, across 47 fresh-context sessions in six phases sharing only on-disk state, with 2{,}162 literature-consultation events. Fault tolerance emerges from redundancy: fresh-context isolation, distributed grounding, and adversarial review catch what any single session misses; pre- and post-pilot stages are fully autonomous, and pilot requires bounded human intervention only at reproduction failures --- operational knowledge curation, not scientific direction. Two paired failure modes --- a pre-architecture baseline and a no-pilot ablation --- isolate structurally enforced numerical confrontation at calibration checkpoints as the operative grounding mechanism. The primitives, characterized failure modes, and quantified intervention pattern lay a foundation for autonomous research in high-stakes scientific domains beyond computational physics.
\end{abstract}

% =====================================================================
\section{Introduction}
\label{sec:intro}

\looseness=-1 AI-for-research has progressed from tool-assistance --- search, summarization, code completion~\citep{taylor2022galactica,chen2021codex,bran2024chemcrow} --- to agentic automation that conducts research with diminishing human oversight~\citep{lu2024aiscientist,yamada2025aiscientist2,schmidgall2025agentlab,miao2025physmaster}. End-to-end automated research has been demonstrated in domains where execution itself provides intrinsic calibration: machine-learning sandboxes where training-loss curves verify methodology~\citep{lu2024aiscientist,yamada2025aiscientist2}, defined computational tasks with established reference values~\citep{miao2025physmaster}, and pre-formulated benchmarks where ground truth is given. Within these calibration-friendly domains, agent reliability has nonetheless been critiqued~\citep{si2024can_novel_research,kambhampati2024position,beel2024aiscientist_critique}: agents over-confidently produce plausible outputs, hallucinate from internal priors, and propagate methodology errors that internal-consistency checks do not catch.

\looseness=-1 As the field pushes toward genuine frontier research --- open direction selection from large literature corpora, novel parameter spaces, underdocumented toolchains where standard recipes do not exist --- execution-as-calibration breaks down structurally. Frontier physical science is the canonical case: physical reasoning is required at every methodology choice, toolchains are often poorly documented, and calibration must come from external literature anchors --- anchors that are, at the frontier, themselves scarce and provisional (\S\ref{sec:discussion}) --- because internal execution provides no ground truth. Without external anchor enforcement, agents default to internal-prior generation, producing confident plausible outputs that cannot be verified by execution alone. The same critiques leveled at simpler automated systems become disqualifying at frontier: an unverifiable result in an ML sandbox is recoverable; an unverifiable result in a physical-science methodology can propagate through to manuscript-level conclusions. This is becoming a blocking problem for the next iteration of automated research.

Real researchers handle this by grounding in literature throughout: from idea conception that surveys what is known and what is open, through methodology calibration that reproduces published references before applying methods to novel computations, to manuscript writing that contextualizes findings against the existing literature. Each function relies on literature in different ways. ``Grounding in literature'' is generic terminology, however --- what does it mean operationally for an autonomous pipeline?

\looseness=-1 We characterize grounding as \emph{structurally enforced literature confrontation at calibration checkpoints --- more than literature availability or consultation count alone}. At calibration checkpoints, the pipeline forces numerical comparison between agent-computed values and published reference values --- not merely citation that references exist. We characterize this distinction empirically via two paired failure modes (\S\ref{sec:grounded}): a pre-architecture baseline run that lacked the topic-selection grounding to reject un-calibratable directions, and a controlled ablation in which curated knowledge and house rules were inherited but the pilot reproduction phase that enforces numerical confrontation was skipped. Together they identify the pilot stage as the architectural element that operationalizes the abstract claim of ``grounded throughout''.

\looseness=-1 The present work runs the full corpus-to-manuscript pipeline on autonomous research-direction selection from open literature, producing a submission-grade artifact in altermagnetic piezomagnetism (companion physics manuscript bundled as-is in \S\ref{appendix:companion}). We choose computational condensed-matter physics as the frontier scope: it has well-characterized symmetry constraints (e.g., point-group constraints on Hall conductivity), reproducible reference frameworks against which calibration is feasible, publicly available software stacks (Quantum~ESPRESSO~\citep{giannozzi2017qe}, Wannier90~\citep{pizzi2020wannier90}, WannierBerri~\citep{tsirkin2021wannierberri}), and an open arXiv corpus base.

\looseness=-1 Our contributions: \textbf{(i)} an end-to-end corpus-to-manuscript pipeline for frontier computational physics --- 47 fresh-context sessions across six phases sharing only on-disk state --- that autonomously produced a submission-grade manuscript with three physics findings; \textbf{(ii)} an operational characterization of ``grounding'' as structurally enforced numerical confrontation at calibration checkpoints (a topic-selection gate plus pilot reproduction), distinct from literature availability or citation; \textbf{(iii)} empirical isolation of this mechanism via two paired failure modes --- a pre-architecture baseline that accepted an un-anchorable topic, and a no-pilot ablation that possessed the anchor value yet never confronted it --- together with a quantified account of the bounded human-intervention pattern (Table~\ref{tab:interventions}); \textbf{(iv)} transferable fault-tolerance primitives (fresh-context isolation, distributed grounding, adversarial review) with 15 documented catch episodes in the pilot stage (Fig.~\ref{fig:anchor}B).

\paragraph{Related work.}\looseness=-1 Two companion studies in the same research program provide context. \citet{huang2026persistent} introduces a persistent knowledge system that lets an LLM agent write down knowledge learned from execution and retrieve it in later sessions, characterizing failure modes in cross-session knowledge transfer. \citet{huang2026scrutiny} establishes grounded autonomous scrutiny at scale: an agent reads a published computational-physics paper, autonomously plans and re-executes the underlying calculations end-to-end, and identifies what does and does not hold; across 111 open-access papers it raises substantive methodological concerns on $\sim$42\%, of which 97.7\% emerge only after execution against a reading-only ceiling of 0.9\%; in depth, on a 2D-MOSFET multiscale simulation paper it produces an unsupervised publishable Comment that revises the headline conclusion. Existing autonomous-research frameworks --- AI Scientist v1/v2~\citep{lu2024aiscientist,yamada2025aiscientist2}, Agent Laboratory~\citep{schmidgall2025agentlab}, PhysMaster~\citep{miao2025physmaster} --- operate in regime~I of \S\ref{sec:discussion}'s taxonomy (calibration supplied by execution feedback or given reference values), and none integrates a first-principles physics toolchain, so a head-to-head run in this domain is not currently constructible; the pre-architecture baseline (\S\ref{sec:grounded}) is the in-domain unscaffolded comparison (framework table: Appendix~\ref{appendix:frameworks}). This work targets regime~III --- the frontier.

% =====================================================================
\section{Pipeline overview}
\label{sec:overview}

\begin{figure*}[!t]
  \centering
  \includegraphics[width=\textwidth]{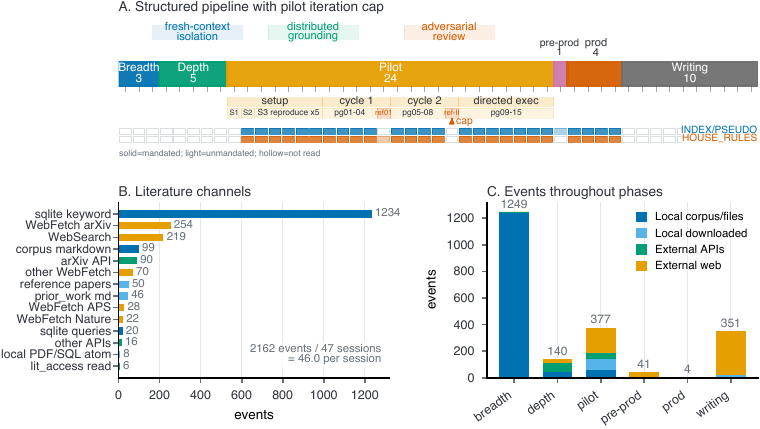}
  \caption{Pipeline architecture and literature footprint. \textbf{(A)}~Six phases run as 47 fresh-context LLM sessions sharing only on-disk state. The pilot lane iterates a computational-gate plus adversarial-review unit; an iteration-cap hyperparameter upgrades the next review to a transition-planning role rather than continuing iteration (details and trade-off in \S\ref{sec:walkthrough}). The bottom band depicts scaffolding access: the \emph{curated knowledge base} (\texttt{INDEX.md}, \texttt{PSEUDOPOTENTIALS.md}) and \emph{house rules} (\texttt{PILOT\_HOUSE\_RULES.md}) are deliberately absent during conception phases (breadth, depth, pilot~S1) and structurally loaded from pilot~S2 onward. \textbf{(B)}~14-channel literature breakdown across the 47 sessions (2{,}162 events). \textbf{(C)}~Per-phase distribution: heavy at breadth (corpus mining), substantial at pilot and writing, near-silent at production.}
  \label{fig:pipeline}
\end{figure*}

\paragraph{Six-phase architecture.} The pipeline runs as six phases (Fig.~\ref{fig:pipeline}A) with an architectural template distinguishing structurally fixed elements from run-specific iteration counts. \emph{Breadth} samples the corpus to extract candidate research themes; the number of breadth sessions is chosen for diversity. \emph{Depth} expands the most promising candidates into research questions with explicit reproduction targets, each session force-generating a new direction not yet committed by previous depth sessions. \emph{Pilot} has fixed structure: program selection (S1), tooling check (S2), $k$ published-reference reproductions, and an iterative gate--review structure with $4{\times}\text{gate} + 1{\times}\text{review}$ cycles bounded by the hyperparameters in \S\ref{sec:pilot}. \emph{Pre-production} (1 session) consolidates pilot outputs into a locked production plan and methodology handoff. \emph{Production} runs as 1 initialization plus adversarial continuations until the agent issues PASS. \emph{Writing} is fixed at three drafting triads plus a polish session ($3{\times}3 + 1 = 10$ sessions). For the canonical run reported here, $n_\mathrm{breadth}=3$, $n_\mathrm{depth}=5$, $k=5$ pilot reproductions, two pilot iteration cycles, and three production continuations.

\paragraph{Two-component scaffolding.}\looseness=-1 Each phase is implemented as one or more LLM sessions, each beginning with a fresh context window: no shared in-memory state, no conversational history. Sessions communicate only through on-disk artifacts. Two architecturally distinct scaffolding components thread through the pipeline. The \emph{curated knowledge base} comprises two files. \verb|INDEX.md| holds curated general knowledge about computational-physics tools and practice --- a single source of truth for ``what can the simulation toolchain do and how'' verified on the local stack, indexing 37 verified workflows, 17 reusable analysis scripts, parameter-variation references, and common operational gotchas; it contains no research-direction-specific content. \verb|PSEUDOPOTENTIALS.md| maps the local pseudopotential library with best-practice selection guidance per workflow. The \emph{house rules} (\verb|PILOT_HOUSE_RULES.md|) enumerate mandatory operational constraints: computational-resource discipline (one heavy job at a time, checkpoint/resume protocols), MUST-do diagnostic steps (Wannier band-fit verification; \texttt{projwfc}/fatband analysis before basis design), and disposition criteria for converged-vs-not verdicts. Both components are loaded only from pilot~S2 onward (Fig.~\ref{fig:pipeline}A bottom band): they are deliberately absent during conception phases (breadth, depth, pilot~S1) to avoid biasing open idea generation toward known methodology. The pipeline thus distinguishes \emph{conception grounding} (open exploration of literature) from \emph{execution grounding} (anchor enforcement via curated knowledge and house rules). Across the 47 sessions, the pipeline performs 2{,}162 unique literature-consultation events distributed across 14 access channels and concentrated in conception and pilot phases (Fig.~\ref{fig:pipeline}B,C).

% =====================================================================
\section{Pipeline walkthrough}
\label{sec:walkthrough}

This section walks through each phase as the agent conceived altermagnetic piezomagnetism from the corpus and ran the pipeline through to manuscript, noting where the architecture's fault-tolerance mechanisms appear in context.

\subsection{Idea conception from corpus}
\label{sec:breadth_depth}

\begin{figure*}[!t]
  \centering
  \includegraphics[width=\textwidth]{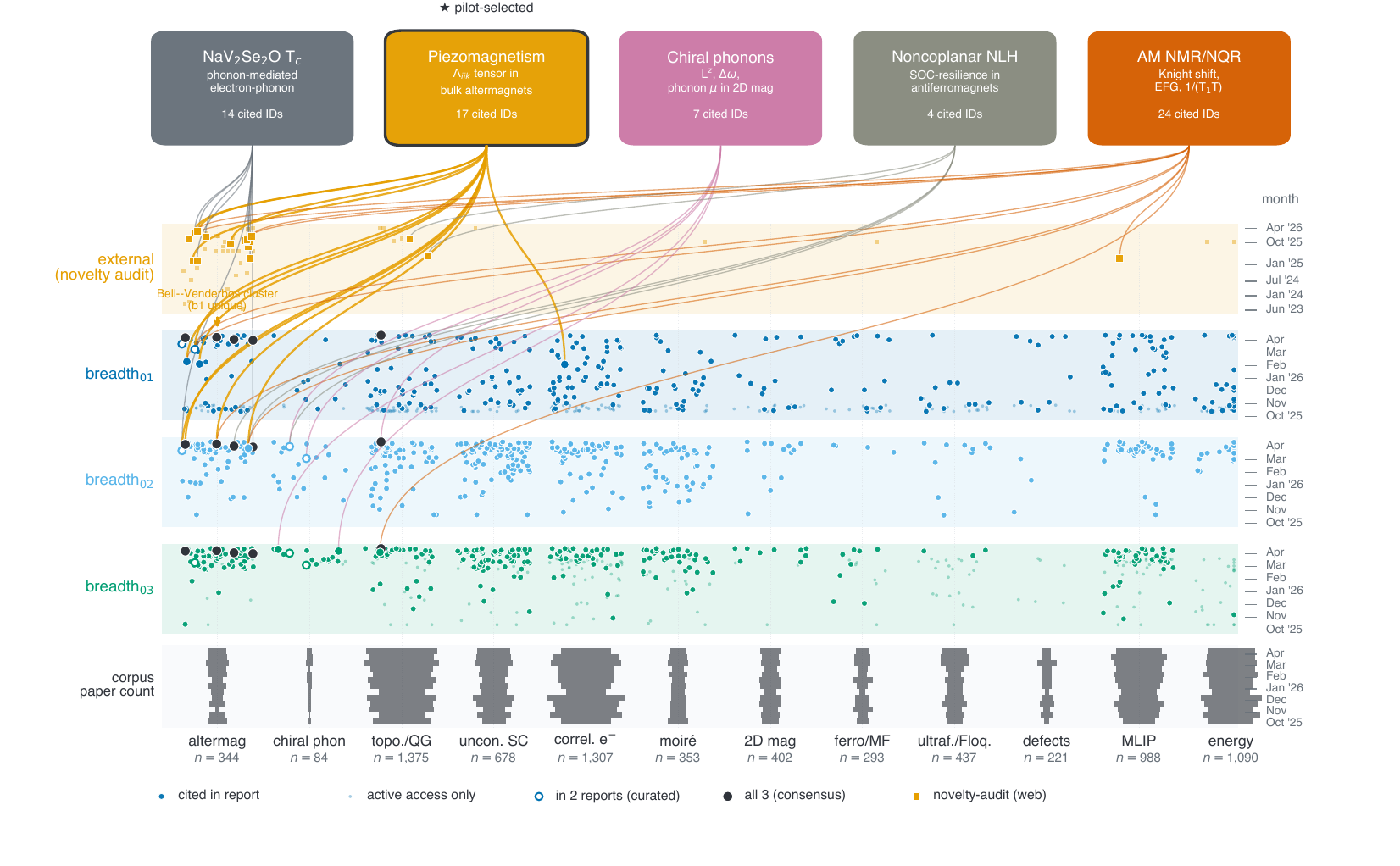}
  \caption{Information flow from corpus to depth programs. Each dot is one arXiv ID; large filled markers are IDs cited in a breadth report or by a depth program, small faint markers are IDs the agent actively retrieved but did not cite. Position: $x$ = regex-classified theme (12 categories cover 99.1\% of breadth-cited and 100\% of depth-cited IDs); $y$ = continuous arXiv submission month. Bands top-to-bottom: \emph{external} (depth-phase novelty audit via OpenAlex / arXiv API / WebFetch / WebSearch, Jun~2023 -- Apr~2026); breadth$_{01}$ (10 parallel \texttt{Explore} sub-agents); breadth$_{02}$ (title walk + 35 SQL queries); breadth$_{03}$ (66 SQL queries). Same arXiv ID has identical $x$ across bands: consensus-core IDs (cited in all three reports) appear as filled black markers stacked vertically; pair-overlap IDs as open rings. Five depth-program boxes at top show committed direction, total cited IDs, and pilot-selected program ($\bigstar$ piezomagnetism); arcs trace each cited ID upward to its destination, with bold arcs for the selected program. Bottom strip: per-theme corpus paper-count histogram in 13 fortnight bins (Oct~2025 -- Apr~2026), absolute scale shared across themes.}
  \label{fig:conception}
\end{figure*}

\looseness=-1 The breadth phase mapped an arXiv corpus of 11{,}083 condensed-matter physics papers across three independent breadth sessions running in parallel (Fig.~\ref{fig:conception}); each session used complementary access strategies. The three reports cite 877 distinct arXiv IDs collectively, with only 27 ($3.1\%$ of the union) appearing in all three --- the breadth agents are 80\% complementary, not redundant. Five depth-phase directions emerged from this distillation, two of which trace primarily to single-channel surfacings --- papers that only one of the three breadth agents identified; the cleanest case, the Bell--Venderbos piezomagnetism cluster surfaced only by breadth$_{01}$'s altermagnetism sub-agent, became the framework of the chosen direction (lineage detail in Appendix~\ref{appendix:lineage}). Empirically, redundancy works because complementarity is high. Depth-phase novelty audits reached external sources --- OpenAlex, the arXiv API, WebFetch, and WebSearch (top external band in Fig.~\ref{fig:conception}) --- both for older reference papers and for in-window papers the breadth agents missed. A depth-phase diversity ledger forced different agent sessions to commit to non-overlapping directions. Table~\ref{tab:candidates} summarizes the five candidates that emerged from this depth phase, with the agent's primary reason and verdict alongside an independent post-hoc human review.

\begin{table}[t]
  \caption{The five candidate research directions evaluated in the depth phase, grouped by source: agent (the pipeline's own composite-score reasoning; verdict labels verbatim from the pilot~S1 program-selection report) and an independent post-hoc human review checking each disposition against current toolchain documentation and published literature.}
  \label{tab:candidates}
  \centering
  \scriptsize
  \setlength{\tabcolsep}{3.5pt}
  \renewcommand{\arraystretch}{1.05}
  \begin{tabular}{@{}p{0.11\linewidth} p{0.27\linewidth} p{0.10\linewidth} p{0.30\linewidth} p{0.11\linewidth}@{}}
    \toprule
    & \multicolumn{2}{c}{Agent (pipeline)} & \multicolumn{2}{c}{Human (post hoc)} \\
    \cmidrule(lr){2-3}\cmidrule(lr){4-5}
    Direction & Reason & Verdict & Reason & Verdict \\
    \midrule
    NaV$_2$Se$_2$O $T_c$ &
    DFPT+EPW $T_c$ for an altermagnetic superconductor at $U$-sensitivity threshold with competing AM ground states within DFT noise. &
    Strong but risky &
    Magnetic-EPW toolchain available only for collinear ferromagnets as of late 2025~\citep{epw2025magnetic}; Allen--Dynes assumes spin-singlet pairing forbidden in altermagnets~\citep{maeland2025altermagnet_pairing}. &
    Drop: methodology development \\
    \textbf{AM piezomagnetism} &
    Strain-induced net magnetization in bulk altermagnets; reproduction targets exist. &
    \textbf{Best composite (selected)} &
    Three independent reproduction anchors verified~\citep{lukashev2008theory,bell2026orbital,ye2026dominant}. &
    \textbf{Selection grounded} \\
    Chiral phonons (2D magnets) &
    Mode-resolved $L^z$ in 2D magnetic insulators; target signal $\sim$1\,meV near a 2--3\,meV noise floor. &
    Noise-floor risk &
    Toolchain~\citep{liu2025chiralphonon_magnetic,huang2026chirality} is 2025--2026 vintage and undertested for 2D magnetic insulators. &
    Defensible drop \\
    Noncoplanar NLH &
    SOC-on/off comparison of the magnetic-geometry channel of intrinsic nonlinear Hall, extending an established study. &
    Safe but incremental &
    Question and toolchain bounded by~\citep{zhu2025nlh_afm}; SOC-resilience generalization is incremental on a mature stack. &
    Drop grounded \\
    AM \mbox{NMR/NQR} &
    Sublattice Knight shifts and EFG fingerprints; pipeline requires QE-CONVERSE + custom Wannier-$\chi(q)$ bridge. &
    Visionary but risky &
    QE-CONVERSE~\citep{ceresoli2025qeconverse} demonstrated only on paramagnetic systems; no altermagnet \emph{ab initio} NMR papers in print as of mid-2026. &
    Drop: methodology development \\
    \bottomrule
  \end{tabular}
\end{table}

\subsection{Pilot stage}
\label{sec:pilot}

\begin{figure*}[!t]
  \centering
  \includegraphics[width=\textwidth]{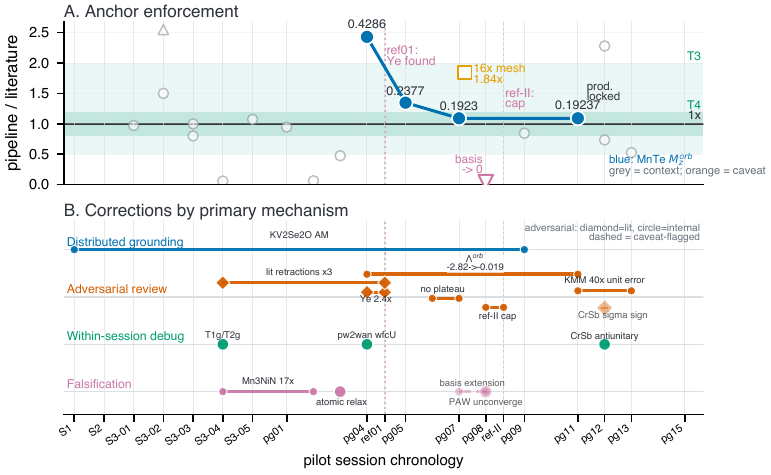}
  \caption{Pilot anchor enforcement and fault tolerance. \textbf{(A)}~MnTe orbital-magnetization $M_z^\mathrm{orb}$ trajectory shown as ratio to Ye~et~al.~\citep{ye2026dominant}'s published value (0.176 $\mu_B$/cell). The first MnTe orbital-magnetization gate's HIGH at 2.44$\times$ is retracted to LOW by the first adversarial review session, whose sub-agent gap-hunt surfaced the published anchor; subsequent gates close the gap step-by-step at the canonical mesh, but the agent itself flagged via NOT PASS that apparent agreement at the canonical mesh was unverified at denser k-mesh. \textbf{(B)}~15 catch episodes during the pilot stage by primary mechanism: distributed grounding (1), adversarial review (7), within-session debug (3), falsification (4); dashed = caveat-flagged.}
  \label{fig:anchor}
\end{figure*}

The pilot phase contains the architectural elements that operationalize the abstract's ``grounded throughout'' claim. It runs as a fixed structure of program selection (S1) and tooling check (S2), followed by reproductions of published references and an iterative gate--review computational sequence. Three fault-tolerance mechanisms appear in this stage: \emph{fresh-context isolation} prevents conclusions from one session being uncritically carried forward by the next; \emph{distributed grounding} redirects a reference missed at one phase to surface at another; \emph{adversarial review} formalizes ``catch what the agent itself overclaimed'' through dedicated reflect, production-continue, and polish sessions prompted to find rather than confirm. In the episode record (Fig.~\ref{fig:anchor}B), \emph{falsification} denotes a dedicated controlled experiment whose outcome refutes a hypothesized cause or fix for a suspect result, as opposed to a re-reading of prior data (adversarial review).

\paragraph{S1 program selection, S2 tooling, S3 reproductions.} S1 selected altermagnetic piezomagnetism from the five depth-phase candidates (Table~\ref{tab:candidates}) and fixed the reproduction set: five published papers targeted as numerical anchors for the pilot reproduction sessions. The independent post-hoc review confirms this selection: piezomagnetism is uniquely well-anchored among the five, with three independent published reproduction targets for the headline observable; the four dropped directions either require methodology-development scope (NaV$_2$Se$_2$O $T_c$, AM NMR/NQR), sit on a young toolchain undertested for the target system (chiral phonons), or are bounded in novelty by an existing study (noncoplanar NLH). This selection gates a deeper point: had the agent committed to a dropped direction, no published reference recipe would exist against which to confront its quantitative findings, and the pipeline could not grade its own output --- topic-selection grounding is the precondition for execution grounding. S2 verified the local toolchain and activated the curated knowledge base and house rules in agent context for subsequent pilot sessions. S3 reproduced each anchor paper individually, comparing computed headline observables against literature values on the pipeline's four-tier verdict scale (T1--T4; defined in Appendix~\ref{appx:multitrack}); each reproduction session was bounded, failure producing documented uncertainty rather than infinite iteration.

\paragraph{Iteration architecture: gate--review cycles.} The pilot phase then operates as bounded iteration: each cycle consists of computational gate sessions followed by an adversarial review session, with two architectural hyperparameters bounding total iteration. \texttt{max\_pilot\_cycle\,=\,$N$} caps the number of normal review sessions before the next review is upgraded to a transition-planning role; \texttt{break\_action} $\in$ \{\texttt{reflect\_II}, \texttt{abort}\} determines whether the cap triggers transition to production with documented uncertainty or termination of the run. The canonical pipeline selected \texttt{max\_pilot\_cycle\,=\,1} and \texttt{break\_action\,=\,reflect\_II}.

The canonical pipeline's actual operation (Table~\ref{tab:cap_tradeoff}, Appendix~\ref{appendix:workflow}; top row): cycle~1 ran four gate sessions plus a review session that retracted an over-confident HIGH verdict (the agent's own confidence rating; HIGH/LOW are the binary outcomes of the composite-score check) to LOW; cycle~2 activated the cap, with the second review upgraded to plan transition into production rather than continuing to iterate. The choice is project-dependent: hypothesis generation justifies a smaller cap, definitive quantitative claims a larger one.

\paragraph{Calibration trajectory: the orbital-magnetization case study.}
\label{sec:morb_saga}

The pilot exercises all three fault-tolerance mechanisms in the calibration trajectory of MnTe orbital magnetization (Fig.~\ref{fig:anchor}A). Altermagnets~\citep{smejkal2022altermagnetism,bhowal2024multipoles,multipoles2025altermagnet,mazin2023altermagnetism} are compensated antiferromagnets with broken time-reversal symmetry; their orbital magnetization $m_\mathrm{orb}$ (distinct from spin magnetization) is the calibration anchor for the production strain-derivative computation. Four moments structure the trajectory.

\textbf{(1) Initial overshoot.}~The agent's first MnTe orbital magnetization produced $|M_z| \approx 0.43$ $\mu_B$/cell, $2.44\times$ above the published reference value, declared HIGH confidence on internal symmetry checks alone (forbidden components $|M_x|, |M_y|$ at noise floor; symmetric-tensor structure verified). The qualitative checks all passed; only numerical comparison against an external published anchor could catch the magnitude error --- exactly the failure mode unscaffolded LLM agents systematically exhibit (\S\ref{sec:intro}).

\textbf{(2) Anchor enforcement in action.}~In the subsequent adversarial review session, opening in fresh context, six sub-agents were deployed prompted to find rather than confirm. One sub-agent's literature gap-hunt surfaced the Ye~2026 reference value (``$M_\mathrm{orb} = 0.176$ $\mu_B$/cell along $z$-axis'')~\citep{ye2026dominant}; the reflect prompt forced the comparison ($2.44\times$, opposite sign), and HIGH was retracted to LOW. The architecture forced the agent to compare against the published reference, not just cite that the reference exists --- anchor enforcement operationally: confrontation, not citation.

\textbf{(3) Convergence verification --- the architecture's NOT PASS verdict.}~The agent calibrated the recipe at one mesh density and matched the Ye reference within 10\%; testing the same recipe at a denser mesh produced 70\% deviation. The agent issued NOT PASS on the convergence point. Rather than continuing to iterate (exceeding the cap), the architecture transitioned to production with the calibrated recipe \emph{plus} documented systematic uncertainty.

\textbf{(4) Trace-substrate verification.}~Computational physics provides a useful substrate for this methodology because every parameter and decision is preserved on disk. A post-completion convergence-ladder test by the human researcher confirmed the divergence direction the agent's NOT PASS had identified. \textbf{The architecture's adversarial review was correct to flag the convergence as unestablished; the manuscript's reduced-scope reporting reflected this honestly --- a pipeline identifying its own convergence limits while producing an anchored attempt is what grounded autonomous research at first iteration looks like.} Fig.~\ref{fig:anchor}B documents 15 catch episodes by primary mechanism across the canonical pilot stage.

\looseness=-1 \textbf{Note added (camera-ready).} After the canonical run completed, follow-up first-principles investigation by the human researcher indicates that the anchor value itself --- Ye et al.'s published MnTe orbital magnetization, $0.176$ $\mu_B$/cell~\citep{ye2026dominant} --- is not a converged observable: the anomalous disentanglement-window sensitivity the pipeline flagged (\S\ref{sec:limits}; Appendix~\ref{app:limits_detail}) is the signature of a basis-truncation systematic, and the published reference value reflects a particular window-and-basis choice rather than a converged orbital magnetization. A full physics account is in preparation as a separate manuscript. This does not alter the architectural conclusions --- every comparison in Figs.~\ref{fig:anchor} and \ref{fig:appendix-multitrack} is anchor-relative, and the architecture behaved correctly under the information available: it enforced confrontation, retracted overconfidence, and refused to certify convergence. It does sharpen what single-anchor calibration provides: consistency with published literature, not truth. A replication-mode pipeline that flags every symptom of an unreliable anchor yet cannot pivot to critiquing the anchor itself is the residual failure mode this episode isolates; diagnostic-triggered switching between replication and critique modes is the corresponding next-iteration architectural direction (\S\ref{sec:discussion}).

\paragraph{Reproduction-failure intervention pattern.}
\label{sec:reproduction_intervention}
A subset of pilot reproductions required human intervention. The pattern is knowledge curation: when the agent fails to converge a reproduction within the bounded session budget, the human researcher debugs offline and adds the resulting general principle to the curated knowledge base. Three substantive additions during the canonical pilot were a \emph{pipeline sanity check} rule (verify basic physical quantities at each calculation step), a \emph{Wannier validation} protocol mandating fatband diagnostic before basis design and \texttt{dis\_froz\_max} plateau scan before any Berry-observable production, and a \emph{Hubbard $U$ non-transferability} clause requiring re-derivation across DFT codes or projector conventions. These are transferable operational principles applicable across computational-physics studies, not study-specific scientific direction. Table~\ref{tab:interventions} quantifies the complete pattern: nine typed in-session events across the canonical run --- four paywall provisioning, four crash/API-recovery green-lights, one compute-status note --- plus the three offline curations and zero scientific-input events. Of the 47 sessions, 44 have archived transcripts; the three breadth sessions --- shared with the baseline run (\S\ref{sec:grounded}) --- entered the canonical run as their on-disk corpus reports and have no archived transcripts, so typed input to them is outside the table's evidence. 38 of the 44, including all ten writing sessions, ran to completion with no typed human input.

\begin{table}[t]
  \caption{Human intervention across the canonical 47-session run, from the archived session transcripts: every typed human message inside a session is counted, excluding per-stage task prompts, one pre-written conditional stage prompt (the below-T3 reproduction-extension patch), and harness-generated records; offline curation counts principles added to the scaffolding after failed reproductions (Appendix~\ref{appendix:kb}). Cf.\ the companion's Author Contributions statement (\S\ref{appendix:companion}).}
  \label{tab:interventions}
  \centering
  \scriptsize
  \setlength{\tabcolsep}{3.5pt}
  \begin{tabular}{@{}p{0.55\linewidth} p{0.11\linewidth} p{0.12\linewidth} p{0.12\linewidth}@{}}
    \toprule
    Intervention type & Count & Phase(s) & Nature \\
    \midrule
    Reference-paper provisioning at prompt-mandated paywall pauses & 4 events, 8 papers & pilot gates & operational \\
    Crash/API-recovery and resume green-lights & 4 & pilot gate, production & operational \\
    Compute-status observation & 1 & pilot S3 & operational \\
    Offline knowledge curation after failed reproductions & 3 principles & pilot & operational \\
    Scientific direction, parameter choice, or interpretation & 0 & --- & --- \\
    \bottomrule
  \end{tabular}
\end{table}

The need for intervention reflects a property of the literature itself: published computational-physics papers vary widely in documentation completeness, from \emph{trivially-replicable} to \emph{frontier-difficult}; \citet{huang2026scrutiny} finds the same spectrum at scale. Reproductions on the difficult end are where human intervention currently remains necessary, with pre-screening and progressive curation as the mitigation. Substantive intervention enters as offline knowledge-base edits between sessions, not as in-context steering (Table~\ref{tab:interventions}). The no-pilot ablation (\S\ref{sec:grounded}), inheriting canonical-pilot knowledge, ran without intervention struggles.

\subsection{Pre-production, production, writing, and outcome}
\label{sec:outcome}

The pre-production session consolidates the pilot recipe (\S\ref{sec:overview}); production-init executes it; three production-continuation sessions run as adversarial review in fresh context. The second adversarial production-continuation session refuted the initial production plan's strain-mode specification, illustrating fresh-context isolation as a robust catch mechanism: the pre-production plan had specified a pure $\varepsilon_{xx}$ scan for CsV$_2$Te$_2$O, but the continuation re-derived the magnetic point group's symmetry constraints, found that the canonical Bell--Venderbos coupling channel is the $B_{1g}$ shear $(\varepsilon_{xx} - \varepsilon_{yy})$, verified via WebFetch~\citep{bell2026orbital}, and switched the protocol before the production endpoints computed. Writing proceeded as three sets of fresh-context drafting sessions (main text with sub-agent reference fetching, supplementary, integration) followed by a final adversarial polish session.

\paragraph{Outcome.} The pipeline produced a submission-grade companion manuscript (\S\ref{appendix:companion}; included as-is, with a usage disclaimer) in $\sim$6 days of wall time, with three substantive findings: \emph{(i)}~the symmetry-permitted anomalous Hall response in MnTe at the deep-valence-band anchor with plateau amplitude $\alpha = 397$ S/cm, the four non-zero N\'eel-vector angles fitting $\alpha\sin(3\varphi)$ within a max residual of $5.8\%$ (envelope band $\pm 15\%$, consistent with the $g$-wave $m'm'm$ signature); \emph{(ii)}~orbital piezomagnetic response in MnTe at the locked recipe with $\Lambda^\mathrm{orb}_{zxx} = -1.96$ $\mu_B$/cell per unit strain and Lukashev asymmetry~\citep{lukashev2008theory} 2.77; and \emph{(iii)}~cross-symmetry-class --- $d$-wave Lieb CsV$_2$Te$_2$O under $B_{1g}$ shear gives $\Lambda_\mathrm{topo} = -24.9$ $\mu_B$/cell per unit shear in the clean-linear regime~\citep{bell2026orbital,radhakrishnan2026topological}. Systematic uncertainties characterized in \S\ref{sec:limits} (and Appendix~\ref{app:limits_detail}).

\subsection{Limitations and mitigation directions}
\label{sec:limits}

The canonical run produced three classes of remaining physics-calibration caveats --- a Mn-3$d$/4$s$ disentanglement-window dependence (which the camera-ready note in \S\ref{sec:pilot} reports extends to the anchor value itself), $k$-mesh convergence in $g$-wave hexagonal Wannier integration, and a 3-point strain extraction --- characterized in Appendix~\ref{app:limits_detail} along with prompt-engineerable mitigations identified for the next pipeline iteration. Each caveat is anchored: identified by the pipeline itself rather than discovered post-hoc, and disclosed in the manuscript as the architecture's NOT~PASS verdict propagating to the published artifact rather than being argued away.

Four scope limitations are structural to this first iteration. (i) One end-to-end research direction: the baseline and ablation are two additional full runs, but on the same direction; cross-direction generalization is untested. (ii) No head-to-head against existing agent frameworks, for the structural reason in \S\ref{sec:intro} (Appendix~\ref{appendix:frameworks}); the pre-architecture baseline is the in-domain comparison. (iii) Multi-trial variance is unquantified: each end-to-end run costs $\approx$6 days of wall time, so trial statistics are deferred. (iv) Full replication is computationally expensive and subject to model-API evolution; the complete prompt set (Appendix~\ref{appendix:scaffolding}) and the on-disk artifacts are the reproducibility surface.

% =====================================================================
\section{What grounded really means: pilot as the grand anchor mechanism}
\label{sec:grounded}

The pipeline accesses literature 2{,}162 times (\S\ref{sec:overview}), but access is not enforced confrontation. Real researchers calibrating methodology do not merely consult published values --- they perform numerical comparison and revise methodology when mismatches emerge. LLM agents pre-trained on internet text default to citing rules and asserting plausibility, not to performing numerical confrontation against literal published values; the pipeline architecture must structurally enforce confrontation at calibration checkpoints. Two paired failure modes characterize this empirically; together they identify the pilot stage as the grand anchor mechanism (run-by-run summary: Table~\ref{tab:threeruns}, Appendix~\ref{appendix:workflow}).

\paragraph{Topic-selection grounding --- the pre-architecture baseline.}
\label{sec:run001}
The pre-architecture baseline run shared the same breadth corpus but lacked a curated knowledge base, a diversity ledger, and the program-selection gate. Three independent depth sessions converged on the same direction --- first-principles phonon thermal Hall in altermagnetic MnTe and CrSb, anchored on a recent experimental observation~\citep{wan2026phonon} --- because breadth-report convergence pulled them there and no diversity-forcing or software-feasibility gate pushed back. The cited theoretical foundations~\citep{bendin2025magnonphonon,hoyer2024magnon,bustamante2026phonon,park2024phonon,weissenhofer2024phonon} had been validated only on toy-model parents differing in crystal symmetry and dimensionality from the 3D NiAs hexagonal $g$-wave altermagnet system. Production wrote $\sim$1{,}200 lines of custom simulation code validated only on the toy-model parents; two post-hoc adversarial review sessions refuted the production headline, and the run terminated without a manuscript. The outputs were un-anchorable by construction. The canonical pipeline's program-selection gate closes this loophole via a software-feasibility clause requiring each stage to be executable by a named tool. In the canonical run, the analogous phonon-thermal-Hall direction was generated as a depth candidate and rejected on this gate. This is \emph{topic-selection grounding}: anchor structure begins at topic acceptance, not at execution.

\paragraph{Execution grounding --- the no-pilot ablation.}
\label{sec:run003}
\looseness=-1 The no-pilot ablation ran the same direction as the canonical pipeline, with the \emph{full curated knowledge and house rules inherited from the canonical post-pilot state}: the curated knowledge base entries (including the \texttt{dis\_froz\_max} selection guide, plateau-scan-mandatory protocol, basis-insufficiency diagnostics, and symmetry-forbidden-component diagnostics) and the house rules (including the Wannier-validation Steps A--F and Checks 1--5 that were expanded during the canonical pilot). The ablation's pre-production literature review independently surfaced Ye~2026 and recorded the reference value (``$M_\mathrm{orb} = 0.176$ $\mu_B$/cell along $z$-axis'')~\citep{ye2026dominant} verbatim; pilot reproduction --- the ablated component --- was skipped, and the run proceeded directly to production.

Despite the published anchor in its own files, the production agent picked a recipe inside the documented collapse tail of the canonical disentanglement-window scan, giving $M_\mathrm{orb,z} = 0.066$ $\mu_B$/cell ($2.7\times$ below the reference). Across all four reflect-continuation cycles, \emph{the comparison ``$0.066$ vs.\ $0.176$'' was never written}. The reference was invoked only via the qualitative ratio $M_\mathrm{orb}/M_\mathrm{spin} \approx 88$, operationally meaningless because $M_\mathrm{spin,z}$ is symmetry-forbidden by $m'm'm$ and identically zero.

\looseness=-1 The defining episode is the dfroz scan run only in the fourth continuation: $M_\mathrm{orb,z}$ collapses from $0.140$ at dfroz $=12.0$ to $0.005$ at dfroz $=13.0$ ($28\times$ across 1\,eV) and $\sigma_{xy}$ flips sign. The agent's verbatim characterization --- ``Plateau is BROAD: 219\% $\sigma_{xy}$ spread, 191\% $M_\mathrm{orb,z}$ spread'' --- is mathematically correct under (max$-$min)/mean but obscures a $28\times$ change with sign flip that is not a ``broad plateau''. The agent then asserted (without testing) that the basis-induced systematic cancels in symmetric finite differences, kept its headlines unrevised, and proceeded to synthesis. The pattern is ``rules cited, not enforced'': with curated rules and Ye's reference value both in its working files, the agent argued individual divergences acceptable case-by-case without performing the literal comparison that would have forced revision. Knowing the published value is not the same as being structurally required to confront against it. This is the gap pilot reproduction's enforced anchor mechanism closes.

\paragraph{Pilot as architectural automatic anchor.}
\label{sec:pilot_anchor}
Both grounding modes --- topic-selection grounding via program selection, execution grounding via pilot reproduction --- are structural mechanisms that operationalize ``literature grounding throughout''. Without them, autonomous pipelines either accept un-anchorable topics or proceed within accumulated rules without enforced numerical confrontation. Crucially, the no-pilot ablation \emph{did} include adversarial production-continuation review across four fresh-context cycles, yet none of these cycles surfaced the literal numerical comparison; this isolates pilot reproduction's structural enforcement --- not adversarial review alone --- as the operative mechanism. The pipeline does not guarantee correctness; it provides anchor structure. Anchored failures (the canonical pipeline's documented systematic uncertainty) become tractable next-iteration directions; ungrounded failures (the ablation's unrevised manuscript) become concealed mystery.

Quantitatively, the canonical run tracked seven published calibration anchors: four end PASS at T4 and three caveat-flagged at their final comparisons, with fails and falsifications confined to intermediate points (Fig.~\ref{fig:appendix-multitrack}); the pilot episode record shows 15 catches across four mechanisms (Fig.~\ref{fig:anchor}B); literature consultation totals 2{,}162 events over 14 channels and 47 sessions (Fig.~\ref{fig:pipeline}B); human intervention totals nine operational in-session events plus three offline curations, none scientific (Table~\ref{tab:interventions}).

% =====================================================================
\section{Discussion}
\label{sec:discussion}

The pipeline mirrors how real research operates: literature throughout, fresh-context adversarial review, knowledge curation across sessions --- behaviors unscaffolded LLM agents skip by default.

\looseness=-1 Anchor-grounded creation is the general, positive contribution: unnecessary in regime~I, where execution itself verifies methodology (the ML sandboxes of \S\ref{sec:intro}); mandatory from regime~II onward (mature domains with consensual, multiply corroborated anchors), where this architecture transfers directly. Regime~III, the frontier, is where this run operated, and it exposes the \emph{frontier anchoring problem}: anchors there are scarce (even in the direction selected as uniquely well-anchored among five, the orbital observable reduced to one published value on the target material, beside a cross-material benchmark) and provisional --- recent, often single results, not yet corroborated or overturned by consensus. Both flavors materialized: an anchor itself unconverged despite maximal apparent provenance --- journal-published, open calculation files (\S\ref{sec:pilot}, Note added) --- and an observable class without field-level convergence (the CrSb gauge spread; the Mazin caveat). Companion work supplies the base rate: substantive methodological concerns on $\sim$42\% of reproduced computational-physics papers~\citep{huang2026scrutiny}. Unreliable and insufficient anchors are the common case at the frontier --- the standing challenge this run makes concrete.

\looseness=-1 The architecture already carries the first-order responses: anchor-density-aware topic selection, a multi-anchor portfolio (Fig.~\ref{fig:appendix-multitrack}), anchor-relative reporting, and NOT~PASS verdicts whose documented sensitivities made post-hoc reattribution of the anchor possible at all. What it deliberately excludes is critique: the debugging discipline ranks the paper itself last among suspects (Appendix~\ref{appendix:scaffolding}) --- a replication prior --- because credible critique is a research program of its own: without execution-grounded scrutiny scaffolding, an agent cannot distinguish literature error from its own unconverged toolchain~\citep{huang2026scrutiny}. The two companion pipelines are complementary halves of one program --- reproduction-anchored creation here, execution-grounded critique at scale there. Both halves are separately demonstrated; their coupling --- diagnostic-triggered escalation from reproduction into scrutiny when anchor-pathology symptoms accumulate as they did here --- is the open problem this pair poses.

\looseness=-1 Two principles emerge. First, \emph{grounded anchor enforcement at calibration checkpoints}: program-selection rejects un-anchorable topics, pilot reproduction forces numerical comparison between agent claims and published anchors; the pre-architecture baseline and no-pilot ablation characterize what fails without each. Second, \emph{fault-tolerant session-breaking adversarial review} (Fig.~\ref{fig:anchor}B): reflect/continue/polish sessions find rather than confirm. The architecture's value is in what it knows it does not know --- NOT~PASS verdicts, documented uncertainty, anchored failures pointing to forward directions. \emph{Can we believe the results} reduces to three operational checks: are agent claims confronted with published anchors; do session-breaking mechanisms catch what single sessions miss; and is each anchor itself corroborated, or single-source and provisional? At the frontier the third check is the demanding one --- which is why anchoring must be read as promising consistency and measured exposure, not truth. When yes on all three, remaining uncertainty becomes identifiable forward direction; when no, concealed mystery.

% =====================================================================
\bibliographystyle{unsrtnat}
\bibliography{references_v2}

\begin{thebibliography}{38}
\providecommand{\natexlab}[1]{#1}
\providecommand{\url}[1]{\texttt{#1}}
\expandafter\ifx\csname urlstyle\endcsname\relax
  \providecommand{\doi}[1]{doi: #1}\else
  \providecommand{\doi}{doi: \begingroup \urlstyle{rm}\Url}\fi

\bibitem[Taylor et~al.(2022)Taylor, Kardas, Cucurull, Scialom,
  et~al.]{taylor2022galactica}
Ross Taylor, Marcin Kardas, Guillem Cucurull, Thomas Scialom, et~al.
\newblock {Galactica}: A large language model for science.
\newblock \emph{arXiv preprint}, 2022.
\newblock arXiv:2211.09085.

\bibitem[Chen et~al.(2021)Chen, Tworek, Jun, et~al.]{chen2021codex}
Mark Chen, Jerry Tworek, Heewoo Jun, et~al.
\newblock Evaluating large language models trained on code.
\newblock \emph{arXiv preprint}, 2021.
\newblock arXiv:2107.03374.

\bibitem[Bran et~al.(2024)Bran, Cox, Schilter, Baldassari, White, and
  Schwaller]{bran2024chemcrow}
Andres~M. Bran, Sam Cox, Oliver Schilter, Carlo Baldassari, Andrew~D. White,
  and Philippe Schwaller.
\newblock {ChemCrow}: Augmenting large-language models with chemistry tools.
\newblock \emph{Nature Machine Intelligence}, 2024.
\newblock arXiv:2304.05376.

\bibitem[Lu et~al.(2024)Lu, Lu, Lange, Foerster, Clune, and
  Ha]{lu2024aiscientist}
Chris Lu, Cong Lu, Robert~Tjarko Lange, Jakob Foerster, Jeff Clune, and David
  Ha.
\newblock The {AI} {{S}}cientist: Towards fully automated open-ended scientific
  discovery.
\newblock \emph{arXiv preprint}, 2024.
\newblock arXiv:2408.06292.

\bibitem[Yamada et~al.(2025)Yamada, Lange, Lu, Hu, Lu, Foerster, Clune, and
  Ha]{yamada2025aiscientist2}
Yutaro Yamada, Robert~Tjarko Lange, Cong Lu, Shengran Hu, Chris Lu, Jakob
  Foerster, Jeff Clune, and David Ha.
\newblock The {AI} {S}cientist-v2: {W}orkshop-level automated scientific
  discovery via agentic tree search.
\newblock \emph{arXiv preprint}, 2025.
\newblock arXiv:2504.08066.

\bibitem[Schmidgall et~al.(2025)]{schmidgall2025agentlab}
Samuel Schmidgall et~al.
\newblock Agent laboratory: Using {LLM} agents as research assistants.
\newblock \emph{arXiv preprint}, 2025.
\newblock arXiv:2501.04227.

\bibitem[Miao et~al.(2025)Miao, Dai, Liu, Tan, et~al.]{miao2025physmaster}
Tingjia Miao, Jiawen Dai, Jingkun Liu, Jinxin Tan, et~al.
\newblock {PhysMaster}: Building an autonomous {AI} physicist for theoretical
  and computational physics research.
\newblock \emph{arXiv preprint}, 2025.
\newblock arXiv:2512.19799.

\bibitem[Si et~al.(2024)Si, Yang, and Hashimoto]{si2024can_novel_research}
Chenglei Si, Diyi Yang, and Tatsunori Hashimoto.
\newblock Can {LLMs} generate novel research ideas? a large-scale human study
  with 100+ {NLP} researchers.
\newblock \emph{arXiv preprint}, 2024.
\newblock arXiv:2409.04109.

\bibitem[Kambhampati et~al.(2024)Kambhampati, Valmeekam, Guan,
  et~al.]{kambhampati2024position}
Subbarao Kambhampati, Karthik Valmeekam, Lin Guan, et~al.
\newblock Position: {LLM}s can't plan, but can help planning in {LLM}-modulo
  frameworks.
\newblock \emph{Proceedings of ICML}, 2024.
\newblock arXiv:2402.01817.

\bibitem[Beel et~al.(2024)]{beel2024aiscientist_critique}
Joeran Beel et~al.
\newblock Evaluating {{Sakana}}'s {AI} {{S}}cientist for autonomous research:
  hidden costs, hidden risks.
\newblock \emph{arXiv preprint}, 2024.
\newblock arXiv:2410.01243.

\bibitem[Giannozzi et~al.(2017)Giannozzi, Andreussi, Brumme, Bunau,
  et~al.]{giannozzi2017qe}
P.~Giannozzi, O.~Andreussi, T.~Brumme, O.~Bunau, et~al.
\newblock Advanced capabilities for materials modelling with {Quantum
  ESPRESSO}.
\newblock \emph{J. Phys.: Condens. Matter}, 29:\penalty0 465901, 2017.
\newblock \doi{10.1088/1361-648X/aa8f79}.

\bibitem[Pizzi et~al.(2020)Pizzi, Vitale, Arita, Bl\"ugel, Freimuth,
  G\'eranton, et~al.]{pizzi2020wannier90}
Giovanni Pizzi, Valerio Vitale, Ryotaro Arita, Stefan Bl\"ugel, Frank Freimuth,
  Guillaume G\'eranton, et~al.
\newblock {Wannier90} as a community code: new features and applications.
\newblock \emph{J. Phys.: Condens. Matter}, 32:\penalty0 165902, 2020.
\newblock \doi{10.1088/1361-648X/ab51ff}.

\bibitem[Tsirkin(2021)]{tsirkin2021wannierberri}
Stepan~S. Tsirkin.
\newblock High performance {Wannier} interpolation of {Berry} curvature and
  related quantities: {WannierBerri} code.
\newblock \emph{npj Comput. Mater.}, 7:\penalty0 33, 2021.
\newblock \doi{10.1038/s41524-021-00498-5}.

\bibitem[Huang(2026{\natexlab{a}})]{huang2026persistent}
Haonan Huang.
\newblock From experiments to expertise: Scientific knowledge consolidation for
  {AI}-driven computational {{physics}}.
\newblock \emph{arXiv preprint}, 2026{\natexlab{a}}.
\newblock arXiv:2603.13191. ICML 2026 AI4Physics Workshop.

\bibitem[Huang(2026{\natexlab{b}})]{huang2026scrutiny}
Haonan Huang.
\newblock Towards grounded autonomous research: an end-to-end {LLM} mini
  research loop on published computational physics.
\newblock \emph{arXiv preprint}, 2026{\natexlab{b}}.
\newblock arXiv:2604.12198. ICML 2026 AI for Science Workshop.

\bibitem[Lihm et~al.(2025)]{epw2025magnetic}
J.-M. Lihm et~al.
\newblock Magnetic extension of the {EPW} code: electron-phonon couplings and
  superconductivity in spin-polarized systems.
\newblock \emph{arXiv preprint}, 2025.
\newblock arXiv:2510.11350.

\bibitem[M{\ae}land et~al.(2025)M{\ae}land, Brekke, and
  Sudb{\o}]{maeland2025altermagnet_pairing}
K.~M{\ae}land, B.~Brekke, and A.~Sudb{\o}.
\newblock Constraints on superconducting pairing in altermagnets.
\newblock \emph{Phys. Rev. B}, 111:\penalty0 014516, 2025.
\newblock \doi{10.1103/PhysRevB.111.014516}.
\newblock arXiv:2408.03999.

\bibitem[Lukashev et~al.(2008)Lukashev, Sabirianov, and
  Belashchenko]{lukashev2008theory}
Pavel Lukashev, Renat~F. Sabirianov, and Kirill Belashchenko.
\newblock Theory of the piezomagnetic effect in {Mn}-based antiperovskites.
\newblock \emph{Phys. Rev. B}, 78:\penalty0 184414, 2008.
\newblock \doi{10.1103/PhysRevB.78.184414}.

\bibitem[Bell and Venderbos(2026)]{bell2026orbital}
Beryl Bell and J\"orn W.~F. Venderbos.
\newblock Orbital piezomagnetic polarizability of pure insulating altermagnets
  in two dimensions.
\newblock \emph{arXiv preprint}, 2026.
\newblock arXiv:2602.10076.

\bibitem[Ye et~al.(2026)Ye, Tenzin, S\l{}awi\'nska, and
  Autieri]{ye2026dominant}
Chao~Chen Ye, Karma Tenzin, Jagoda S\l{}awi\'nska, and Carmine Autieri.
\newblock Dominant orbital magnetization in the prototypical altermagnet
  {MnTe}.
\newblock \emph{Phys. Rev. B}, 113:\penalty0 014413, 2026.
\newblock \doi{10.1103/PhysRevB.113.014413}.
\newblock arXiv:2505.08675.

\bibitem[Liu et~al.(2025)]{liu2025chiralphonon_magnetic}
Y.~Liu et~al.
\newblock {Ab initio} framework of electron-phonon coupling for chiral phonons
  with giant phonon magnetic moments in magnetic materials.
\newblock \emph{Phys. Rev. Lett.}, 135:\penalty0 256701, 2025.
\newblock arXiv:2503.10160.

\bibitem[Huang et~al.(2026)Huang, Kusuno, Hashimoto, Juraschek, Kusunose, and
  Satoh]{huang2026chirality}
Z.~Huang, A.~Kusuno, M.~Hashimoto, D.~M. Juraschek, H.~Kusunose, and T.~Satoh.
\newblock Quantifying chirality of phonons.
\newblock \emph{arXiv preprint}, 2026.
\newblock arXiv:2604.10231.

\bibitem[Zhu et~al.(2025)Zhu, Li, Chen, Yu, and Liu]{zhu2025nlh_afm}
Y.~Zhu, J.~Li, Z.~Chen, S.~Yu, and Q.~Liu.
\newblock Spin-orbit-coupling-resilient nonlinear {{Hall}} response in
  non-coplanar antiferromagnets.
\newblock \emph{Nat. Commun.}, 16:\penalty0 4882, 2025.
\newblock arXiv:2406.03738.

\bibitem[Ceresoli et~al.(2025)]{ceresoli2025qeconverse}
D.~Ceresoli et~al.
\newblock {QE-CONVERSE}: a non-perturbative orbital-magnetization module for
  {Quantum~ESPRESSO}.
\newblock \emph{arXiv preprint}, 2025.
\newblock arXiv:2503.04664.

\bibitem[{\v{S}}mejkal et~al.(2022){\v{S}}mejkal, Sinova, and
  Jungwirth]{smejkal2022altermagnetism}
Libor {\v{S}}mejkal, Jairo Sinova, and Tom{\'a}{\v{s}} Jungwirth.
\newblock Beyond conventional ferromagnetism and antiferromagnetism: {A} phase
  with nonrelativistic spin and crystal rotation symmetry.
\newblock \emph{Phys. Rev. X}, 12:\penalty0 031042, 2022.
\newblock \doi{10.1103/PhysRevX.12.031042}.

\bibitem[Bhowal and Spaldin(2024)]{bhowal2024multipoles}
S.~Bhowal and N.~A. Spaldin.
\newblock Magnetoelectric classification of skyrmions and altermagnets.
\newblock \emph{Phys. Rev. X}, 14:\penalty0 011019, 2024.
\newblock \doi{10.1103/PhysRevX.14.011019}.

\bibitem[Hayami et~al.(2025)]{multipoles2025altermagnet}
S.~Hayami et~al.
\newblock Multipole framework for altermagnetic order parameters.
\newblock \emph{arXiv preprint}, 2025.
\newblock arXiv:2512.17587.

\bibitem[Mazin(2023)]{mazin2023altermagnetism}
I.~I. Mazin.
\newblock Altermagnetism in {MnTe}: origin, predicted manifestations, and
  routes to detwinning.
\newblock \emph{Phys. Rev. B}, 107:\penalty0 L100418, 2023.
\newblock \doi{10.1103/PhysRevB.107.L100418}.
\newblock arXiv:2301.08573.

\bibitem[Radhakrishnan et~al.(2026)Radhakrishnan, Bell, Ortix, and
  Venderbos]{radhakrishnan2026topological}
H.~Radhakrishnan, B.~Bell, C.~Ortix, and J.~W.~F. Venderbos.
\newblock {Topological piezomagnetic effect in two-dimensional Dirac quadrupole
  altermagnets}.
\newblock \emph{arXiv preprint}, 2026.
\newblock arXiv:2602.05894.

\bibitem[Wan et~al.(2026)]{wan2026phonon}
Y.~Wan et~al.
\newblock Anomalous phonon thermal {Hall} effect in altermagnetic {MnTe}.
\newblock \emph{arXiv preprint}, 2026.
\newblock arXiv:2604.03183.

\bibitem[Bendin et~al.(2025)Bendin, Mook, Mertig, and
  Neumann]{bendin2025magnonphonon}
S.~Bendin, A.~Mook, I.~Mertig, and B.~Neumann.
\newblock Magnon-phonon hybridization in {2D} square-lattice altermagnets.
\newblock \emph{arXiv preprint}, 2025.
\newblock arXiv:2511.08357.

\bibitem[Hoyer et~al.(2024)Hoyer, Jaeschke-Ubiergo, Ahn, {\v S}mejkal, and
  Mook]{hoyer2024magnon}
M.~Hoyer, R.~Jaeschke-Ubiergo, J.~Ahn, L.~{\v S}mejkal, and A.~Mook.
\newblock Magnon {Hall} effect in altermagnets: minimal model and material
  predictions.
\newblock \emph{arXiv preprint}, 2024.
\newblock arXiv:2405.05090.

\bibitem[Bustamante-Lopez et~al.(2026)Bustamante-Lopez, Brehm, and
  Juraschek]{bustamante2026phonon}
B.~Bustamante-Lopez, B.~Brehm, and D.~M. Juraschek.
\newblock Atomistic framework for phonon angular momentum and {Hall} effects.
\newblock \emph{arXiv preprint}, 2026.
\newblock arXiv:2604.01899.

\bibitem[Park et~al.(2024)Park, Nagaosa, and Oh]{park2024phonon}
S.~Park, N.~Nagaosa, and J.~Oh.
\newblock Phonon thermal {Hall} effect in $\alpha$-{RuCl}$_3$ from first
  principles.
\newblock \emph{arXiv preprint}, 2024.
\newblock arXiv:2407.00660.

\bibitem[Wei{\ss}enhofer et~al.(2024)]{weissenhofer2024phonon}
M.~Wei{\ss}enhofer et~al.
\newblock Magnon-phonon thermal {Hall} effect in bcc iron from first
  principles.
\newblock \emph{arXiv preprint}, 2024.
\newblock arXiv:2411.03879.

\bibitem[Lopez et~al.(2012)Lopez, Vanderbilt, Thonhauser, and
  Souza]{lopez2012wannier}
M.~G. Lopez, David Vanderbilt, T.~Thonhauser, and Ivo Souza.
\newblock {Wannier}-based calculation of the orbital magnetization in crystals.
\newblock \emph{Phys. Rev. B}, 85:\penalty0 014435, 2012.
\newblock \doi{10.1103/PhysRevB.85.014435}.

\bibitem[Khodas et~al.(2026)Khodas, Mu, Mazin, and
  Belashchenko]{khodas2026tuning}
M.~Khodas, Sai Mu, I.~I. Mazin, and K.~D. Belashchenko.
\newblock Tuning of altermagnetism by strain.
\newblock \emph{Phys. Rev. B}, 113:\penalty0 104422, 2026.
\newblock \doi{10.1103/PhysRevB.113.104422}.
\newblock arXiv:2506.06257.

\bibitem[Marzari et~al.(2012)Marzari, Mostofi, Yates, Souza, and
  Vanderbilt]{marzari2012wannier}
Nicola Marzari, Arash~A. Mostofi, Jonathan~R. Yates, Ivo Souza, and David
  Vanderbilt.
\newblock Maximally localized {Wannier} functions: {T}heory and applications.
\newblock \emph{Rev. Mod. Phys.}, 84:\penalty0 1419--1475, 2012.
\newblock \doi{10.1103/RevModPhys.84.1419}.

\end{thebibliography}

\clearpage
\appendix
\begin{center}
{\Large\bfseries Appendix}
\end{center}
\vspace{1em}

\section{Multi-anchor convergence detail and tier scheme}
\label{appx:multitrack}

\textbf{Tier scheme as operated} (defined in the reproduction template and pilot-gate prompt, Appendix~\ref{appendix:scaffolding}): T4 = all mandatory targets within their per-reference tolerances, with $\leq$20\% the default where the reference specifies none (gate sessions map T4 onto a HIGH confidence rating and PASS verdicts); T3 = qualitative match (signs and trends correct, magnitudes offset); T2 = partial reproduction; T1 = failure. Mandatory tolerances are set per reference paper (from 0.1\% on a lattice constant to factor-2 on a near-band-edge Hall conductivity), so $\leq$20\% is the canonical default, not a universal cutoff.

\begin{figure}[h]
  \centering
  \includegraphics[width=0.95\linewidth]{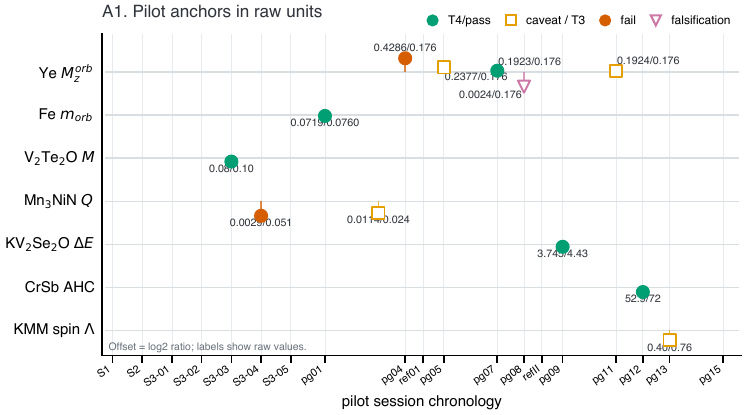}
  \caption{Pilot-stage anchor comparisons in raw units. Each row tracks one published reference (Ye orbital magnetization~\citep{ye2026dominant}, Lopez Fe orbital magnetization~\citep{lopez2012wannier}, V$_2$Te$_2$O magnetization, Mn$_3$NiN multipole, KV$_2$Se$_2$O level splitting, CrSb anomalous Hall, Khodas--Mu--Mazin spin piezomagnetic coefficient~\citep{khodas2026tuning}); points are pipeline-vs-literature ratios on a $\log_2$ axis at the session in which the comparison was performed. Markers: solid green (T4 PASS: within the target's mandatory tolerance, $\leq$20\% default), open square (T3, caveat-flagged), solid orange (fail), pink triangle (falsification). The Ye row encapsulates the orbital-magnetization trajectory of \S\ref{sec:morb_saga}.}
  \label{fig:appendix-multitrack}
\end{figure}

The figure complements Fig.~\ref{fig:anchor}A by showing the full multi-anchor structure across all calibration targets. Each anchor entered the pipeline as a published reference value; the agent's pipeline-output value at each session is divided by the anchor and plotted on the log-ratio axis. PASS marks T4 agreement (within the target's mandatory tolerance, $\leq$20\% default; see \S\ref{sec:pilot} and the tier scheme above); caveat-flagged points are order-of-magnitude consistent (T3) but miss the strict match. The visual pattern is the empirical content of ``grounded throughout'': every calibration checkpoint is anchored to a specific published value, and every disagreement enters the lessons that downstream production and writing inherit.

% =====================================================================
\section{Detailed limitations and mitigation directions}
\label{app:limits_detail}

The canonical pipeline produced a publication-grade artifact with three classes of remaining caveats, each pointing to a community-relevant question or prompt-engineerable mitigation.

\paragraph{Small disentanglement-window plateau.} The MnTe orbital magnetization shows a $\sim$0.1\,eV plateau in the disentanglement upper bound, narrower than the $\mathcal{O}(1$\,eV$)$ insensitivity typical for Wannier frozen windows~\citep{marzari2012wannier,pizzi2020wannier90}. A post-pipeline \emph{Step~C} fatband identifies the mechanism as a steeply-ramping Mn-$4s$ contribution outside the window. Disentanglement-window selection for orbital observables in this class warrants plateau-scan verification.

\paragraph{$k$-mesh sensitivity.} A 70\% shift between $12^3$ and $16^3$ NSCF mesh densities at fixed window, above the sensitivity expected from Fe-class benchmarks~\citep{lopez2012wannier}. The agent's NOT PASS verdict was correct (\S\ref{sec:morb_saga}); a post-pipeline convergence-ladder test confirmed the direction. Per-$k$ decomposition localizes the residual at BZ corners $A$, $H$ (band near-degeneracies). Altermagnetic Wannier orbital magnetization appears to require denser convergence than benchmarks suggest.

\paragraph{3-point strain protocol.} The orbital piezomagnetic coefficient was extracted from three matched-gauge symmetric-FD points; Lukashev asymmetry~\citep{lukashev2008theory} 2.77 from three points exactly fits a 3-parameter parabola without independent verification of functional form. Prompt-engineerable: the reflect prompt can mandate ``if Lukashev $> 2$, add additional strain points''.

A \emph{Step~C} fatband diagnostic, though mandatory in the house rules, was not run in either canonical pilot or ablation --- to be enforced via reflect prompt in the next iteration.

% =====================================================================
\section{Pipeline scaffolding --- prompts and house rules}
\label{appendix:scaffolding}

Each session in the pipeline is driven by a structured prompt template that encodes the agent's workflow: inputs to read, numbered task steps, outputs to produce, and iteration logic. Prompts are versioned during development as failure modes become understood (e.g., the pilot-gate prompt has nine versions across the canonical pilot's evolution; we include the latest). We provide the five most architecturally consequential prompts verbatim below, and a structural summary of the remainder; the full prompt set is available in the archived project repository (doi:~\texttt{10.5281/zenodo.21126996}).

\subsection*{C.1 Breadth prompt}
\textit{Source: \texttt{breadth\_prompt\_v2.md}; used in 3 canonical sessions.}
\begin{lstlisting}[basicstyle=\tiny\ttfamily]
# Task: Research Landscape Mapping

## The mission

You are working in a large project that aims to produce a first-principles computational physics paper suitable for peer review at a respectable journal. The standard is novelty, rigor, and thoroughness throughout. The final paper must make a clear and substantial scientific contribution -- not incremental, not a routine calculation, not a minor variation on existing work.

## Your task

This is an **extensive reading task**. You read the recent arXiv literature in condensed-matter and adjacent physics and produce a map of the research landscape: what people are working on, where the activity is, what directions look scientifically promising. A later agent will use your map, together with its own deep reading, to conceive specific research directions.

You do not propose research ideas. Your output is a map. Ideation is the next stage's job.

## Reading material

**Primary source -- local arXiv corpus (read-only):** the most recent 6 months of papers from six arXiv categories: cond-mat.mtrl-sci, cond-mat.mes-hall, cond-mat.supr-con, cond-mat.str-el, physics.chem-ph, physics.comp-ph. ~11k papers total.

- `<PROJECT>/corpus/title_list.md` -- every paper in the corpus, one line per paper: `[arxiv_id] (category, date) Title -- journal_ref?`. **Cat this entire file into context at session start; do not skim or sample.** Journal references are present for ~10% of papers (only those whose authors manually added a publication record on arXiv); absence does not mean unpublished.
- `corpus/corpus.db` -- SQLite. Query for abstracts and full metadata. Columns include `arxiv_id`, `title`, `abstract`, `primary_category`, `all_categories`, `submission_date`, `doi`, `journal_ref`. Use it freely.
- `corpus/markdown/<arxiv-id>.md` -- full paper text in markdown, converted from PDF. Read selectively for papers worth understanding deeply.
- `corpus/pdfs/<arxiv-id>.pdf` -- original PDF. Read directly if a markdown conversion looks garbled or you need to inspect figures the markdown cannot represent.
- `corpus/latex/<arxiv-id>.tar.gz` -- original LaTeX source for ~78% of papers. If markdown is garbled or you need clean equations/figure captions, extract on demand into your own workspace: `tar -xzf corpus/latex/<id>.tar.gz -C <your_workspace>/latex_unpacked/<id>/`. Do not extract into the corpus directory.

**Secondary source -- full internet access (on demand):** for anything beyond the corpus -- older literature, specific paper lookup, broader context. Before reaching for the generic web search tool, read `<PROJECT>/knowledge/literature_access.md`. It documents the structured APIs available (arXiv all-time, Semantic Scholar, OpenAlex, Crossref, Elsevier TDM with key) with code examples. Structured APIs are higher precision than web search; use them first.

**Filesystem discipline:** the corpus directory and knowledge directory are **read-only** for you. Write only under your own working directory. Any extracted LaTeX, downloaded PDFs, or intermediate files go in subdirectories of your workspace.

## What to look for

The corpus is diverse. Good landscape entries include:

- **Dominant themes**: directions where many papers cluster.
- **Emerging bursts**: recent activity that has not yet consolidated.
- **Underrepresented but substantive directions**: scattered high-quality work that has not converged into a visible theme. Low competition.
- **Gaps**: experimental observations without clean theoretical understanding, theoretical predictions awaiting experimental test, contradictory findings, methodology shifts mid-corpus.
- **Cross-pollinations**: ideas that appear across subfield boundaries.

A first-principles computational study could make a contribution in many of these. The map should surface areas where such a contribution is plausible -- not restricted to papers that already used computation. Experimental papers, reviews, and theory papers are all fair game.

## Deliverable

A single file `breadth_report.md` in your working directory. Suggested structure:

1. **Themes** -- dominant directions, with representative arXiv IDs and a sense of activity level.
2. **Emerging patterns** -- new bursts, methodology shifts, anomalies.
3. **Underrepresented directions** -- scattered but substantive, low competition.
4. **Gaps** -- where a first-principles study could contribute.
5. **Candidate directions for deep-dive** -- a ranked shortlist (5--10) of directions worth dedicated investigation. Rank by scientific potential, not by heat. Heat and novelty both matter; explicitly note which drives each ranking.

Use arXiv IDs inline as evidence. Do not fabricate citations. For each claim, distinguish between "I read N papers in this area" and "I inferred this from abstracts alone" -- mark your epistemic state per theme.

If you find yourself writing "someone should study X" or "we propose Y" -- stop and rewrite. Say "there is a gap around X" or "direction Y appears unexplored." You are mapping, not ideating.

## Begin

Read the resources, develop your own reading strategy, and produce `breadth_report.md`. **No time limit, no token limit -- do not save tokens, this stage matters.** Your effort here feeds directly into every downstream stage; the ideation agents can only work with what you surface. Take the time this task deserves.
\end{lstlisting}

\subsection*{C.2 Depth prompt}
\textit{Source: \texttt{depth\_prompt\_v3.md}; used in 5 canonical sessions.}
\begin{lstlisting}[basicstyle=\tiny\ttfamily]
# Task: Research Program Formulation

## The mission

You are working in a large project that aims to produce a first-principles computational physics paper suitable for peer review at a respectable journal. The final paper must make a clear and substantial scientific contribution.

## Your task

Read landscape maps from prior agents, dive deep into a promising direction, read substantial full-text literature, and formulate a research program -- a main question with sub-questions and a fallback -- that could become a publishable paper. A pilot agent will later audit feasibility; your job is to deliver a program strong enough to survive that audit.

## Input

- `breadth_reports/` -- landscape maps from prior agents. Read all in full.
- `chosen_topics.md` -- directions already chosen by prior depth agents. **Read before committing.** Your direction must differ substantially from any listed. After committing, append your direction (one line) to this file.
- **Corpus** (read-only, ~11k papers, 6 months): `corpus/title_list.md` (cat in full), `corpus/corpus.db` (SQLite), `corpus/markdown/<id>.md`, `corpus/pdfs/<id>.pdf`. Path: `<PROJECT>/corpus/`.
- **Internet access** for novelty audit. Read `<PROJECT>/knowledge/literature_access.md` first for structured APIs (arXiv all-time, Semantic Scholar, OpenAlex, Crossref, Elsevier TDM).
- Write only in your own workspace.

## Constraints

**Hardware**: 12 cores, 48 GB RAM, ~24h production compute. **Toolchain**: DFT, DFPT, TDDFT, GW, Wannier interpolation, ballistic transport -- standard first-principles methods assumed available. Do not exceed this ecosystem.

**Software feasibility gate.** For each simulation stage in the proposed pipeline (not analysis or plotting), you must be able to name a specific existing open-source tool or package that performs it. If a stage cannot be done by a named tool and would require writing custom simulation code -- even if the underlying formulas are well-documented -- that stage is methodology development, not pipeline execution. The scientific contribution should be in the physical insight, not the computational method.

## Workflow

This is an **iterative** process, not a linear one. You explore multiple candidates and screen them before committing deeply to one.

**Stage 1 -- Candidate generation.** Read breadth reports and `chosen_topics.md`. Identify 2--3 candidate research directions (different from any in `chosen_topics.md`). For each candidate, write a brief sketch: one-paragraph description + the computational pipeline as a list of stages. Do not deep-read literature yet.

**Stage 2 -- Feasibility screen.** For each candidate's pipeline, apply two gates:

- **Hardware gate:** Can each stage plausibly complete within 24h on 12 cores / 48 GB? Are the system sizes reasonable?
- **Software gate:** For each simulation stage, name the specific open-source tool that performs it. If any stage has no named tool and would require custom simulation code (constructing Hamiltonians, implementing solvers, deriving formulas from papers), the candidate fails.

If at least one candidate passes both gates -> proceed to Stage 3 with the best one.

If all candidates fail -> **return to Stage 1** with fresh candidates. You may return up to twice (3 rounds total, screening 6--9 candidates). After 3 rounds with no viable candidate, simplify the most promising candidate from all rounds until it passes both gates.

**Stage 3 -- Commit + deep reading.** Select the best candidate that passed both gates. Append your direction to `chosen_topics.md`. Now dive deep: read full-text papers, develop main question + sub-questions + fallback.

**Stage 4 -- Novelty audit.** The corpus covers only 6 months. Audit every question against the broader literature using external APIs. For each question: has it been done? If partially, what is your contribution beyond it? If the direction is completely done, try the next passing candidate from Stage 2 -- or return to Stage 1 if none remain.

**Stage 5 -- Finalize.** Only a program that has passed both feasibility screen and novelty audit is finalized.

## Deliverable

A single `research_program.md`:

1. **Candidates considered** -- the 2--3 directions you screened, with pipeline sketch and feasibility outcome for each. Include dropped candidates and why they were dropped.
2. **Direction and rationale** -- which candidate you committed to, why, how it differs from `chosen_topics.md` entries.
3. **Literature foundation** -- full-text papers read, 1--2 lines each on what you took from them.
4. **Main research question** -- specific, first-principles-addressable.
5. **Sub-questions** (2--4) -- concrete enough for a pilot agent to estimate cost.
6. **Fallback path** -- if the main question fails, what salvageable story remains?
7. **Prior work + novelty** -- per question: relevant prior work (URL, title, year, 1-line takeaway) + differentiation.
8. **Computational feasibility** -- per pipeline stage: named tool, has it been demonstrated on a similar system, hardware estimate.
9. **Expected deliverable** -- what the paper would argue, one paragraph.
10. **Computational footprint** -- classes of calculation, system sizes, 24h/48GB plausibility.

## Begin

Read `chosen_topics.md`, then breadth reports. Start with candidate generation and feasibility screening -- do NOT deep-read literature until a candidate has passed both gates. **No time limit, no token limit.** The program you conceive here determines the ceiling of impact for the entire project.
\end{lstlisting}

\subsection*{C.3 Pilot gate prompt (latest of 9 versions)}
\textit{Source: \texttt{pilot/pilot\_gate\_prompt} latest version; used in 15 canonical pilot-gate sessions.}
\begin{lstlisting}[basicstyle=\tiny\ttfamily]
# Task: Pilot Gate -- Iterative Pipeline Confidence Assessment

## The mission

You are working in a large project that aims to produce a first-principles computational physics paper suitable for peer review at a respectable journal. Prior agents have reproduced 5 published papers to validate the computational pipeline. Your job is to assess whether the pipeline is ready for production -- and if not, to close the most critical gap.

## Your task

**Depth over breadth.** One thing fully investigated to quantitative agreement with literature is worth infinitely more than five things explored qualitatively. Do not speculate about what "might work." Pick one gap, close it, and prove it with numbers.

The production pipeline can only be trusted on a new material if every computational step has been demonstrated to give quantitative agreement (?20%) with published literature on a physically similar system. A step that was only smoke-tested on a toy system, or that showed >2x disagreement with unresolved diagnosis, is not validated -- it is a guess.

## Input

**Prior work (mandatory reads -- read ALL before writing your verdict):**

- `session_3_reproduce_*/` -- all 5 reproduction sessions. Read each `reproduction_report.md`, `paper_verdict.json`, and skim each `WORKLOG.md` for methodology lessons and unresolved issues.
- `pilot_gate_*/` -- any previous pilot gate sessions (if they exist). Read each `pilot_gate_report.md` and `pilot_gate_verdict.md`. Pay attention to what was attempted, what worked, and what didn't.
- `pilot_reflect_*/` -- any previous pilot reflect sessions (if they exist). These contain literature searches, published reference values, and recommendations from a fresh-perspective review. **Pay close attention -- these identify specific papers and numbers you should reproduce or compare against.**
- `pilot_reflect_II_*/` -- any previous pilot reflect II sessions (if they exist). These contain paper figure analyses, gap analyses, and **planned pilot_gate tasks with specific deliverables.** **Pilot reflect II sessions supercede everything else regarding tasks. If a reflect II report assigns you a specific task, execute it instead of picking your own gap in Phase 2.**
- `session_1_prior_work/program_selection.md` -- the production program. What pipeline stages does production require? What are the headline claims?

**Local environment:**

- `runs/run_002/knowledge/TOOL_REPORT.md` -- central tool registry. **Read first.**
- `runs/run_002/knowledge/` -- per-tool docs, custom-script docs.
- `runs/run_002/tool_examples/` -- verified examples per tool.
- `runs/run_002/.venv/` -- Python venv: `source <PROJECT>/runs/run_002/.venv/bin/activate`

**Global knowledge (mandatory reads):**

- `<PROJECT>/knowledge/PSEUDOPOTENTIALS.md`
- `<PROJECT>/knowledge/INDEX.md` -- contains critical computational gotchas and lessons learned. **Read the sections relevant to your pipeline tools and calculation types carefully before computing anything. Failure in doing so will result in time wasted and garbage produced.**
- `<PROJECT>/knowledge/literature_access.md`
- `PILOT_HOUSE_RULES.md` in your workspace -- **read before any planning and computation. Obey EVERYTHING inside. Even if any required procedure increases work significantly, DO IT (like U scan and Wannier band verification). Failure of doing so is regarded complete failure of the whole task since skipping required works and gates will certainly result in garbage data. Especially: 1. after starting a calculation, DO ABSOLUTELY NOTHING (except for checking RAM/core usage) UNTIL IT IS DONE. You should have only ONE monitor to see when the job stops. Do not spawn multiple monitors only to confuse yourself. One job at a time, so no more than one monitor. 2. DO NOT KILL AND RESTART AN SCF RUN UNLESS YOU ARE SURE THE NUMBER STAYS EXACTLY THE SAME FOR >10 ITERATIONS OR HAS OSCILLATED MULTIPLE TIMES. Evidence has shown that you like to grow impatient and kill scf multiple times, resulting in no result. 3. For Wannier-based calculations: convergence warnings are not cosmetic. If disentanglement reports "criteria not satisfied", downstream results are unreliable -- do not proceed. Read corresponding PILOT_HOUSE_RULES parts and INDEX lessons on frozen window before setting up any Wannier chain.**

**Prior reproduction artifacts:** `session_3_reproduce_*`, `pilot_gate_*`, and `pilot_reflect_*` workspaces are read-only. **Copy any useful intermediate results (converged SCF, Wannier checkpoints, U scan data) into your own workspace.** But only inherit critically -- if a prior result didn't match literature quantitatively, don't blindly reuse it. Improve on it, or simply start from scratch. For QE, you can copy `outdir/` and use `restart_mode='restart'` to save SCF time when the cell and magnetic configuration are compatible.

**Compute:** 12 cores, 48 GB RAM. See `TOOL_REPORT.md` for invocation commands.

**Filesystem:** Everything outside your workspace is read-only. Your workspace as `pilot_gate_NN/` (ls the pilot directory to determine the next number). All output goes in your workspace only. You may install additional Python packages if necessary -- document in worklog.

**Pseudopotentials:** Virtually all QE-compatible pseudopotential libraries are available locally (see `PSEUDOPOTENTIALS.md` for paths). You should not need to download PPs -- just pick the right ones from what's available. If you genuinely need a PP that isn't local, you may search online and download it.

## Workflow

### Phase 0: Create your workspace as `pilot_gate_NN/` (ls the pilot directory to determine the next number). **All output goes in your workspace only**.

### Phase 1: Confidence assessment and verdict (DO THIS FIRST -- before any calculation)

Read all prior work. **Think critically and independently, as a skeptical referee would. THIS REFLECTION PART IS EXTREMELY IMPORTANT BECAUSE IT IS THE FOUNDATION OF THE WHOLE TASK: TAKE YOUR TIME, DIG DEEP AND THINK CAREFULLY.** Prior agents might neglect important details from raw data, their chain-of-thought reasoning may also be wrong, and their attributions for why a result deviated may be incorrect. Can the results between different runs really compared? Did they compare unconverged with converged results? Do multiple factors change at the same time, contributing to the difference? Dig into raw data. Can you find any difference of parameters that the previous agent neglected, which could tell a different story? Moreover, they lack the global view you now have across all 5 reproductions and any prior pilot gate sessions. Do not take their diagnoses at face value -- verify them against your own physics understanding and against raw data.

During Phase 1, you may search online and quickly fetch paper abstracts or values to inform your verdict. But this is a preliminary scan -- it will miss supplementary details, datasets, and methodological nuances. Do not assume you have read a paper because you skimmed its arXiv page. The full reading with published version, supplementary, and dataset is a separate mandatory step in Phase 2.5, and cannot be skipped even if you already fetched the arXiv version.

Imagine the production paper is published alongside all the reproduce and pilot_gate session data as supplementary material. A referee asks: "You couldn't reproduce [X] quantitatively -- how do I know your production results on a new material aren't noise?" That is the standard you are defending against.

For each pipeline stage that production requires, assess your confidence:

- **HIGH**: this step was tested on a physically similar system AND gave quantitative agreement (?20%) with published literature. The recipe (parameters, pseudopotentials, convergence settings) is documented and transferable.
- **MEDIUM**: this step was tested but only qualitatively (correct signs/trends, wrong magnitudes), OR on a system with different physics than production targets, OR with untested assumptions about why it didn't match quantitatively.
- **LOW**: this step was smoke-tested only (simple system, not production-relevant physics), OR showed >2x disagreement with diagnosis that is speculation rather than verified explanation.
- **NONE**: this step was never executed in any session.

**For each non-HIGH stage, do the following analysis:**

1. **Identify the gap precisely.** What quantitative target was missed, and by how much?
2. **Review prior diagnoses.** What did the reproduce/pilot_gate agents say caused the deviation? Do you agree with their reasoning? What alternative explanations exist?
3. **List concrete possibilities** for closing the gap -- with your assessment of how likely each is to work and why.
4. **REQUIRED: Search online.** For each gap, search documentation, forums, mailing lists, tutorials, examples, and GitHub issues for the relevant tools and physics. Many problems in computational physics are well-known community issues with documented solutions. **Do this search before deciding what to work on -- it may completely change your assessment of which gap is tractable.** Judge the authority of sources yourself (official docs > tutorials > forum posts > random blogs).
5. **Assess tractability.** Is there a clearly promising path, or has this been extensively attempted across multiple sessions without convergence? Do not go down rabbit holes -- if prior sessions have made extensive, well-reasoned attempts and the gap remains, it may be genuinely hard. Prioritize gaps where you see a concrete, evidence-based path to T4 over gaps that are high-impact but have no clear solution.

Write `pilot_gate_verdict.md` with:
1. A confidence table (one row per production pipeline stage).
2. For each non-HIGH stage: the full analysis above (gap, prior diagnoses, your assessment, online research findings, possibilities, tractability).
3. Overall verdict: **PASS** (all production-critical stages at HIGH) or **NOT PASS** (list gaps ranked by combined impact x tractability).

**If PASS, stop here.** Write `pilot_gate_report.md` summarizing why and finalize.

**If NOT PASS, proceed to Phase 2.**

### Phase 2: Pick ONE targeted project

From your NOT PASS gaps, pick **the single gap that best combines high impact on production confidence with a clear, evidence-based path to resolution.** Not two gaps. Not three. One.

You have two options:

**Option A -- Reproduce one new paper.** Download and reproduce a published paper that directly exercises the gap. Same workflow as `session_3_reproduce` sessions (reading_summary, WORKLOG, comparison figures, verdict). Use literature_access.md for paper retrieval. Pick a paper where the answer is known so you can verify quantitatively.

**Option B -- Enhance a previous result or run a targeted test.** This could include: convergence studies (k-mesh, cutoff), pseudopotential comparison, U scan to match a known observable, SOC on/off comparison, or any other controlled test that produces a quantitative number compared to published literature.

### Phase 2.5: MANDATORY -- Read the reference paper before any calculation

**This applies to BOTH Option A and Option B.** Any time you are comparing your result against a published number, you MUST read the source paper thoroughly first. A number without its context is meaningless -- you need to understand what code, what parameters, what conventions produced that number before you can meaningfully compare.

**Step 1: Get the published version.** If the paper has a journal-published version, you MUST use it. An arXiv preprint is NEVER the published version, even if the content appears identical -- arXiv lacks supplementary materials, datasets, and final methodological details, and using it wastes sessions and computation. Do not attempt to bypass paywalls or download published PDFs yourself. Stop, create a folder in your workspace (e.g., `reference_papers/`), and ask the user: give them the DOI, tell them you need the main text PDF, supplementary PDF, and any linked code or dataset. Wait for "continue."

**Step 2: Read the full paper -- main text AND supplementary AND dataset.** Not just the abstract. Not just one number from a table. Read the computational details section word by word. Read the supplementary material. If a dataset or code repository is linked, examine **every file** -- input files, scripts, output logs, intermediate products. Not just files for our code -- files for ANY code the paper used. The paper may use a different DFT code, but every parameter in their input files has a physical meaning that maps to our pipeline. Extract ALL of them.

**Step 3: Replicate the recipe.** The goal is **faithful replication of the paper's computational setup**, translated to our pipeline. Do not fixate on the seemingly biggest discrepancy or contribution and selectively pick which parameters to match and which to dismiss. Very often, some details you ignore at the beginning turn out to be important. So, pay attention to everything. Specifically:

- Map every parameter from the paper's input files to our pipeline equivalent. If the paper uses a different code, read the documentation for BOTH codes to understand what each parameter does and find the correct equivalent. Do not assume parameter names are transferable across codes -- follow the references the paper cites for each methodological choice.
- For parameters not explicitly stated in the main text (e.g., number of bands, energy windows, convergence thresholds), analyze thoroughly the dataset's input files, output logs, and figures. Inspect the paper's figures carefully -- band structure plots, convergence plots, and DOS figures encode implicit parameter information.
- If a parameter cannot be determined from any source or it is genuinely not comparable due to different software, note it explicitly and make a physically motivated choice. Document your reasoning.

Write `parameter_comparison.md` (MANDATORY deliverable): every parameter from the paper, the equivalent in our pipeline, and how you will match it. This is not a document for deciding what matters -- it is a checklist for confirming you have matched everything.

**After writing `parameter_comparison.md`, STOP.** Re-read `PILOT_HOUSE_RULES.md` and the relevant sections of `INDEX.md`. Check: does your parameter comparison cover everything the rules require? Have you dismissed any parameter as "can't map" or "probably not important" that a rule actually addresses? Prior sessions have written parameter comparisons, dismissed the most important difference, and wasted entire sessions debugging the wrong thing. If a parameter differs and you haven't matched it, it is a candidate problem -- not something to explain away.

**Step 4: Validate by comparing intermediate products against the reference.** If the dataset includes intermediate outputs (Wannier spreads, band structures, convergence logs), compare yours against theirs at each pipeline stage. If intermediates don't match, stop and think deeply about physics -- what is physically different? Fix the upstream cause before proceeding. A final number produced without intermediate validation is not trustworthy. If your final result does not match the paper, you have two tools: (1) compare intermediates more deeply to locate where the pipeline diverges, and (2) go back to `parameter_comparison.md` -- every parameter difference you listed but did not match is a candidate cause. The answer is almost always in the parameters you haven't matched yet.

**Step 5: Write `reading_summary.md`.** Summarize the paper's physics, your complete parameter mapping, and your replication plan. The final verdict must come from running the full pipeline end-to-end with our own tools -- but intermediate comparisons against the dataset are essential for debugging.

**Pseudopotential choice is often the most impactful variable.** Before starting any calculation, check what pseudopotential files are available locally (in the PP directories documented in PSEUDOPOTENTIALS.md). Are there alternative choices (different libraries, different core-valence partitions, PAW vs NC) that might give better results for your specific system? If a prior reproduction showed systematic deviation, pseudopotential mismatch is a prime suspect.

**The goal is to reach T4 (?20% agreement with literature) on this one thing.** Write a concrete plan in WORKLOG before starting: what is the central question, what quantitative target will you compare against, what parameter mapping ensures an apples-to-apples comparison, and what does success look like.

### Phase 3: Execute

**Before launching any calculation, STOP and re-read `PILOT_HOUSE_RULES.md` in full AND the relevant sections of `INDEX.md` for the tools you are about to use.** Do this now, not from memory -- hours of reading and planning have passed since session start. Every rule and check in these documents was written because ignoring it has resulted in garbage data and days of wasted computation. This is not hypothetical: prior sessions have lost 10+ hours to violations of rules that were read at session start and forgotten by calculation time. Re-read. **OBEY `PILOT_HOUSE_RULES.md` IN FULL, especially all those checks and gotchas.** Then compute.

**Validate on a single point before any sweep.** If the paper or test sweeps a parameter, run ONE representative point first. Verify it is physically sound and within reasonable range of the target before investing hours in the full sweep.

**Think critically about how many points you need.** The goal is to verify quantitative agreement, not to produce pretty curves.

**Prioritize the quantitative target.** One key value, properly converged and matching literature, is the entire deliverable. Everything else is secondary.

For each calculation:
- Process check before launching (house rules ?1), log to WORKLOG.
- Launch with proper parallelism. Block fully.
- On success: extract numbers, compare to literature.
- On failure: diagnose. Failures are where pipeline gaps are discovered.

### Debugging discipline

When a result deviates, suspect in this order:
1. Your custom code (conventions, signs, units, prefactors).
2. Your input parameters (compare against paper parameter by parameter).
3. Tool version or configuration mismatch.
4. The paper itself (last resort -- only after exhausting 1-3).

Use the internet for physics conventions and tool documentation (house rules ?10).

### Custom script discipline

Custom scripts require ?3 independent validations on published reference values before their outputs enter the verdict. Document validations in worklog.

### Phase 4: Updated assessment (APPEND, do not overwrite)

After completing your targeted project, write `pilot_gate_report.md`. This is a NEW document -- do NOT overwrite or modify `pilot_gate_verdict.md` from Phase 1. The Phase 1 verdict is the "before" snapshot; the report is the "after." Both must be preserved for provenance.

`pilot_gate_report.md` should contain:
- What gap you targeted and why.
- What you did (briefly).
- Quantitative result vs literature.
- Updated confidence assessment for the stage you worked on -- did the gap close?
- Remaining gaps (for the next pilot_gate iteration, if needed).

**Re-evaluate honestly.** If your targeted project did NOT close the gap (result still >20% off, or revealed a deeper problem), say so. Do not inflate confidence because you spent time.

## Time discipline

- **Start:** Record a timestamp in WORKLOG.md.
- **At 6 hours:** Assess progress. If the quantitative target is not yet matched, reduce scope -- focus exclusively on getting one clean number.
- **At 8 hours:** Stop launching new calculations. Write up.
- **At 10 hours:** Hard stop. Finalize all deliverables with whatever data exists.

## Deliverables

| File | Content |
|------|---------|
| `pilot_gate_verdict.md` | Confidence table + gap analysis + PASS/NOT PASS (written FIRST, before calculations) |
| `WORKLOG.md` | Continuous log |
| `parameter_comparison.md` | Every parameter from reference paper/dataset mapped to our pipeline -- checklist for confirming complete replication, not selective assessment (MANDATORY) |
| `reading_summary.md` | Paper physics, complete parameter mapping, replication plan (MANDATORY) |
| `pilot_gate_report.md` | What was done, quantitative results, updated confidence, remaining gaps (written AFTER calculations, preserving Phase 1 verdict separately) |
| Calculation artifacts | In subdirectories within your workspace |
| `comparison_*.png` | Quantitative comparison figures |

## Begin

First, `ls` the pilot directory to determine your session number. Create `pilot_gate_NN/`. Read `PILOT_HOUSE_RULES.md`, then read ALL prior `session_3_reproduce_*/reproduction_report.md`, `pilot_gate_*/pilot_gate_report.md`, and `pilot_reflect_*/reflect_report.md` (if any exist), and 
`pilot_reflect_II_*/reflect_II_report.md` (if any exist). Then read `program_selection.md`. Write your confidence verdict BEFORE touching any calculation. The production stage inherits your assessment -- if you say HIGH and it's not, production builds on a false foundation.
\end{lstlisting}

\subsection*{C.4 Pilot reflect prompt}
\textit{Source: \texttt{pilot/pilot\_reflect\_prompt\_1.md}; used in 1 normal pilot-reflect session.}
\begin{lstlisting}[basicstyle=\tiny\ttfamily]
# Task: Pilot Reflect -- Analyze, Search Literature, Find Problems

You do not run any calculations. You do not modify anything outside your own workspace. Create `pilot_reflect_NN/` (ls pilot/ for the next number). Your only deliverable is `reflect_report.md`.

## Read

Everything in `pilot/`: all `reproduction_report.md`, `pilot_gate_report.md`, `pilot_gate_verdict.md`, prior `reflect_report.md`. Also `session_1_prior_work/program_selection.md`.

## Your primary task

Identify every pipeline step production requires. For each step, identify the key physical quantity. For each quantity, answer:

**What do we get? What does the literature say?**

Search broadly -- arXiv, Google Scholar, journal sites. Find published theoretical values for the same quantity. Prefer same material, then same material family, then same physics on any material. Find experiments too. Read the papers our agents tried to reproduce -- and find papers they didn't try. Follow-up papers, competing calculations, reviews.

For each quantity where our pipeline disagrees with a paper we tried to reproduce, diagnose:

- **Is it a literature problem?** Does the broader literature (other theory groups, experiments) support our value or the paper's? If multiple independent sources agree with us, the paper may be an outlier -- find a better reference to validate against.
- **Is it a methodology problem?** Did our agents use shortcuts, skip steps the paper did, use different pseudopotentials without testing alternatives? Check unit conventions, sign conventions, normalization (per atom vs per cell, Ry vs eV, etc.). If the disagreement is a constant factor, think hard about whether it's a convention mismatch before concluding it's physics.
- **Is it a pipeline problem?** If results are noisy rather than a consistent offset, that points to convergence or implementation issues, not literature disagreement.
- **Is it a code problem?** Check if any custom code, patches, or agent-derived formulas were used. Verify them independently.

Also scrutinize quantities where agents declared agreement -- is the agreement real, or was it against an internal benchmark rather than an independent published number?

## Your secondary task

Based on your analysis, issue a verdict.

**PASS:** every production-critical pipeline step has ?20% agreement with at least one published theoretical value on a physically similar system. Theory is the primary source for quantitative comparison. Experimental agreement provides qualitative confidence (correct sign, right order of magnitude) but does not substitute for theory-vs-theory quantitative benchmarking.

**NOT PASS (expected outcome):** provide a task list for the next pilot_gate round. For each task: the gap, a specific paper to reproduce or test to run (with arXiv ID and the specific number to match), and a decision tree -- under what conditions can this gap be considered closed? Under what conditions should we stop trying and accept the uncertainty?

**PARTIAL PASS (last resort):** issue only if a specific pipeline step has been extensively attacked (?3 independent theoretical references attempted, convention/unit issues ruled out, pipeline systematics investigated) and still cannot be reproduced. In that case, recommend dropping that specific claim from production scope. This is a scope reduction, not a free pass.

## Write `reflect_report.md`

Per-step analysis with our values, literature values (cited), diagnosis. Task list if NOT PASS. Keep it concrete -- every recommendation should point to a specific number in a specific paper.

If a paper is behind a paywall or download fails, pause and ask the user to download it. Tell them the paper ID, where to put the PDF, and say "continue."
\end{lstlisting}

\subsection*{C.5 Pilot reflect\_II prompt (iteration-cap activation)}
\textit{Source: \texttt{pilot/pilot\_reflect\_II\_prompt.md}; used at the iteration-cap activation event.}
\begin{lstlisting}[basicstyle=\tiny\ttfamily]
# Task: Pilot Reflect II -- Self-Validation Strategy and Production Plan

You do not run any calculations. You do not modify anything outside your own workspace. Create `pilot_reflect_II_NN/` (ls pilot/ for the next number).

## Context

Multiple rounds of pilot_gate and pilot_reflect sessions have been run. The goal is no longer to fit a specific number from a specific paper. The goal is to understand the physics behind your numbers, identify what is still missing to support them, and plan the remaining work before production.

## Step 1: Study the published papers -- figures AND physics logic

Find every paper that was reproduced or referenced across all reproduce and pilot_gate sessions. Use the published version with SI if available (check `reference_papers/` folders in prior pilot_gate sessions first; those have publisher PDFs, SI, and datasets). For each paper:

**Physics logic chain.** How does the paper explain the physical mechanism behind its headline quantity? What produces this quantity at the microscopic level? What orbitals, interactions, or symmetry-breaking mechanisms are responsible? How does the paper connect the mechanism to its computed result? Quote the key sentences verbatim -- the ones where the authors explain WHY a quantity has the value it does, not just WHAT the value is. These logic chains tell you what physical understanding is required to interpret and defend a result.

**Figures.** Look at every figure in main text AND SI. For each figure:
- What physical quantity is displayed, and what is it plotted against (energy, k-point, strain, doping)?
- What does this dependence reveal about the mechanism? (e.g., a k-resolved plot shows WHERE in the Brillouin zone a quantity originates; an energy-resolved plot shows WHICH bands contribute)
- How does this figure support or diagnose the headline quantity? Could it reveal why a calculation succeeded or failed?
- Is this figure related to any quantity our pipeline computes?

Write `paper_figure_analysis.md`: paper by paper, figure by figure, logic chain by logic chain. This is a deliverable -- not a mental exercise.

## Step 2: Review everything in pilot/

Read all `reproduction_report.md`, `pilot_gate_report.md`, `pilot_gate_verdict.md`, `pilot_reflect_*/reflect_report.md`, and `session_1_prior_work/program_selection.md`.

**Think critically.** Prior agents may lack the global view you now have. They may have overlooked details or drawn wrong conclusions. Any doubt -- check raw data (calculation outputs, logs, numbers) rather than trusting a prior agent's verdict or claim.

What numbers do we have? What recipes work? What sensitivities have we discovered? What remains unresolved? Write a concise summary grounded in raw data.

## Step 3: Gap analysis -- what do we need for a paper?

Imagine writing a paper from our existing pilot data. Compare against the papers analyzed in Step 1:

- For each paper's physics logic chain: does our work include the supporting quantities and analyses that chain requires? If a paper explains a quantity through a specific mechanism and supports that explanation with a specific type of analysis (e.g., k-resolved decomposition of a Brillouin-zone integral, orbital-projected band structure, energy-dependent integration), have we done the equivalent? If not, flag it.
- What figures and analyses did those papers include that we have NOT done? For each: essential (paper rejected without it), important (strengthens argument), or nice-to-have?
- Are there quantities we computed but never validated with a diagnostic that would reveal whether the physics is correct?
- Are there parameter sensitivities we discovered but never explained -- we know THAT something is sensitive, but not WHY? Is there an underlying physical quantity whose analysis would reveal the mechanism?

## Step 4: Plan remaining pilot_gate sessions

Write a concrete plan for closing the gaps from Step 3. The goal is **understanding**, not expanding scope. Prioritize:

1. **Big gaps that block production.** If production requires a pipeline step that has never been validated, or a quantity that was computed but is physically unsupported, this must be addressed. New calculations are justified here.
2. **Supporting analyses for existing results.** We have numbers from prior sessions -- what analyses and figures would make them defensible? The goal is to understand why the pipeline gave the results it did, not to add more calculations for their own sake. If we discovered a sensitivity, compute the diagnostic that explains it.

Each planned session should be:
- A specific computation or analysis with a clear deliverable (a figure, a convergence study, a validated recipe)
- Sized to fit within a 6-10h pilot_gate session (one task per session)
- Ordered by priority (production blockers first)

For each planned session, specify:
- What to compute and why (tied to a specific gap from Step 3)
- What figure(s) to produce
- What analysis to write -- load and inspect every figure you produce, check if the physics makes sense, write what you learned. A figure without interpretation is not a deliverable.
- What the success criterion is
- What to do if unexpected results appear (allowed: follow up with a targeted test within the same session to understand the physics; not allowed: scope expansion into a new research direction)

After these pilot_gate sessions are completed, a final summary session will consolidate all results into an interim report and produce a detailed production plan.

## Deliverables

| File | Content |
|------|---------|
| `paper_figure_analysis.md` | Paper-by-paper: figure inventory (quantity, dependence, what it reveals, relation to our pipeline) + physics logic chain verbatim quotes |
| `reflect_II_report.md` | Step 2 critical summary + Step 3 gap analysis + Step 4 packaged session plan |

If a paper is behind a paywall or download fails, pause and ask the user to download it. Tell them the paper ID, where to put the PDF, and say "continue."
\end{lstlisting}

\subsection*{C.6 Other prompts: structural summary}
The remaining ten canonical prompts share a common structure: explicit numbered task steps, mandatory inputs to read, outputs to produce, and iteration / disposition criteria. Brief summary:

\begin{itemize}[leftmargin=*,nosep,topsep=2pt]
\item \textbf{Pilot S1 program selection} (\texttt{pilot/session\_1\_prior\_work\_prompt.md}, 1 session): read the five depth research-program reports; rate each against composite criteria (novelty / computational tractability / tool readiness / physics depth, each 1--5); apply software-feasibility gate; select one direction; nominate reproduction targets for S3; output \texttt{program\_selection.md} + \texttt{reproduction\_targets.md}.
\item \textbf{Pilot S2 tooling readiness} (\texttt{pilot/session\_2\_tooling\_prompt.md}, 1 session): verify the local toolchain operates against a small reference test for each pipeline stage; load curated knowledge and house rules into agent context; output \texttt{TOOL\_REPORT.md}.
\item \textbf{Pilot S3 reproduction template} (\texttt{pilot/session\_3\_reproduce\_template\_3.md}, instantiated 5$\times$): given a target paper, reproduce its central headline observable; compare against the published value on the T1--T4 tier scale (T4 = all mandatory targets within their per-reference tolerances, $\leq$20\% default; see \S\ref{sec:pilot}); output \texttt{reproduction\_report.md} with explicit numerical comparison.
\item \textbf{Pre-production} (\texttt{pre\_production/pre\_production\_prompt.md}, 1 session): consolidate pilot artifacts into a locked production plan; final literature review; tentative manuscript story; methodology handoff.
\item \textbf{Production initial} (\texttt{production/production\_prompt.md}, 1 session): execute the locked recipe; numbered tier-priority task list; honest disposition (PASS / NOT~PASS) per task; do not exceed scope.
\item \textbf{Production continuation} (\texttt{production/production\_continue\_prompt\_2.md}, 3 sessions): operate as adversarial review of the prior production session in fresh context; re-derive symmetry constraints; re-fetch references when needed; correct or refine.
\item \textbf{Writing 1A} (\texttt{paper/pre\_write\_1A\_prompt.md}, 3 sessions): main-text drafting with sub-agent fetching of $\sim$202 reference abstracts; produce numbered manuscript sections.
\item \textbf{Writing 1B} (\texttt{paper/pre\_write\_1B\_prompt.md}, 3 sessions): supplementary-material drafting; figure-caption development; tables.
\item \textbf{Writing full} (\texttt{paper/full\_write\_prompt.md}, 3 sessions): integration of 1A + 1B; consistency check; final figure placement.
\item \textbf{Polish} (\texttt{paper/polish\_prompt.md}, 1 session): final adversarial review on the integrated draft; structural and prose pass; verify cross-references.
\end{itemize}

\subsection*{C.7 House rules}
\textit{Source: \texttt{pilot/PILOT\_HOUSE\_RULES.md}; loaded into agent context from pilot~S2 onward.}
\lstinputlisting[basicstyle=\tiny\ttfamily]{prompts/PILOT_HOUSE_RULES.md}

\subsection*{C.8 Curated knowledge base}
The curated knowledge base \texttt{INDEX.md} (470 lines) and \texttt{PSEUDOPOTENTIALS.md} are large and primarily methodology-reference content rather than research-direction-specific. We summarize their scope and evolution in Appendix~\ref{appendix:kb}. Full text and the complete prompt set above are available in the archived project repository (doi:~\texttt{10.5281/zenodo.21126996}).

% =====================================================================
\section{Per-task workflow observations}
\label{appendix:workflow}

\begin{table}[t]
  \caption{Architectural trade-off for the pilot iteration cap (referenced from \S\ref{sec:pilot}). The selected setting favors maximum output per fixed compute budget: a calibrated manuscript artifact with characterized caveats, at the cost of remaining convergence work being deferred to the next iteration.}
  \label{tab:cap_tradeoff}
  \centering
  \small
  \setlength{\tabcolsep}{4pt}
  \begin{tabular}{@{}p{0.34\linewidth} p{0.28\linewidth} p{0.13\linewidth} p{0.13\linewidth}@{}}
    \toprule
    Choice & Output & Compute & Future basis \\
    \midrule
    \texttt{max\_pilot\_cycle\,=\,1}, \texttt{reflect\_II} (selected) & Anchored manuscript with characterized caveats & Moderate & High \\
    \texttt{max\_pilot\_cycle\,$\geq\!2$} & Same artifact, tighter uncertainty bounds & High & Higher \\
    \texttt{break\_action\,=\,abort} & None (no manuscript, no calibration retained) & Low & Low \\
    \bottomrule
  \end{tabular}
\end{table}

\begin{table*}[t]
  \caption{The canonical run and its two paired failure modes. Both failure modes ran the full production/continuation machinery; they differ only in which grounding structures were present. (Referenced from \S\ref{sec:grounded}.)}
  \label{tab:threeruns}
  \centering
  \scriptsize
  \setlength{\tabcolsep}{4pt}
  \renewcommand{\arraystretch}{1.1}
  \begin{tabular}{@{}p{0.105\textwidth} p{0.10\textwidth} p{0.135\textwidth} p{0.125\textwidth} p{0.16\textwidth} p{0.26\textwidth}@{}}
    \toprule
    Run & Topic-selection gate & Curated knowledge + house rules & Pilot reproduction & Literal anchor confrontation & Outcome \\
    \midrule
    Canonical & enforced & loaded from pilot S2 & 5 anchors + gate--review cycles & enforced (Figs.~\ref{fig:anchor}A, \ref{fig:appendix-multitrack}) & anchored manuscript with documented caveats; 15 catch episodes \\
    Pre-architecture baseline & absent & absent & absent & impossible (no published recipe exists for the accepted topic) & $\sim$1{,}200 lines of toy-validated custom code; headline refuted by two post-hoc reviewers; no manuscript \\
    No-pilot ablation & direction inherited & inherited (post-pilot state) & \textbf{skipped} & never performed (``0.066 vs.\ 0.176'' never written across four reflect cycles) & unrevised headlines from a recipe inside the documented collapse tail \\
    \bottomrule
  \end{tabular}
\end{table*}

\begin{figure*}[!htbp]
  \centering
  \includegraphics[width=\textwidth]{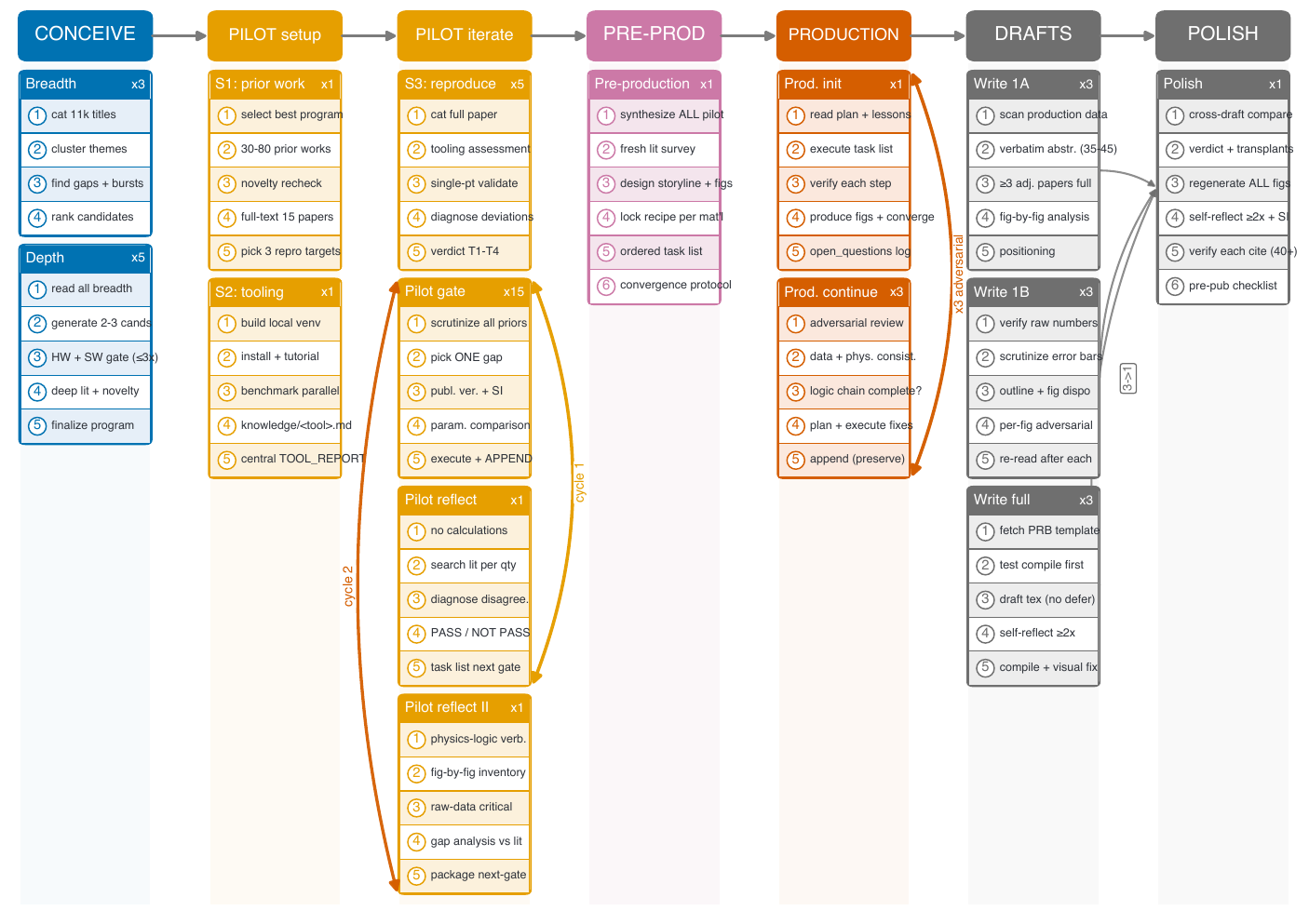}
  \caption{%
\textbf{Per-phase prompt-mandated workflow of the canonical run pipeline.}
Pipeline phases run left to right across seven columns; within each column,
task cards stack vertically; within each card, prompt-mandated workflow steps
stack top-to-bottom with adjacent step boxes touching.
Title-bar ``$\times N$'' denotes the number of canonical sessions of that
task type. Iteration arrows: cycle~1 (orange) is the
\textit{Pilot gate}\,$\leftrightarrow$\,\textit{Pilot reflect} loop on the right
of \textsc{Pilot iterate}; cycle~2 (vermillion) is the
\textit{Pilot gate}\,$\leftrightarrow$\,\textit{Pilot reflect~II} loop on the
left. \textsc{Production}: \textit{Prod.\ init}\,$\leftrightarrow$\,\textit{Prod.\ continue}
$\times 3$ adversarial loop (vermillion). \textsc{Drafts}: \textit{Write~1A},
\textit{Write~1B}, \textit{Write~full} each $\times 3$ in parallel; \textsc{Polish}
merges three drafts (3$\rightarrow$1 fan-in). Step content is paraphrased verbatim
from canonical prompts; full prompt sources are available in the archived project repository (doi:~\texttt{10.5281/zenodo.21126996}).}
  \label{fig:phase-workflow}
\end{figure*}

Two structural observations summarize the prompt set in Appendix~\ref{appendix:scaffolding}. First, most tasks are predominantly linear in execution: numbered steps with explicit inputs and outputs, the agent following the workflow without branching. Iteration concentrates in the pilot phase (gate--reflect cycles) and at production-continuation boundaries (adversarial review); the writing phase has a polish-cycle iteration but is otherwise linear. Second, sub-agent spawning is concentrated in two places: breadth (10 parallel \texttt{Explore} sub-agents fanning out by theme) and pilot reflect (multiple gap-hunt sub-agents prompted to find rather than confirm). The structured nature of each task --- explicit numbered steps with mandatory inputs and outputs, disposition criteria specifying when to iterate vs.\ when to transition --- is what makes the architecture transferable across LLM substrates and physical-science subdomains. Figure~\ref{fig:phase-workflow} renders the prompt-mandated workflow per phase across the canonical run.

% =====================================================================
\section{Corpus and 12-theme categorization}
\label{appendix:corpus}

\paragraph{Corpus characterization.} The arXiv corpus comprises 11{,}083 papers from six categories (\texttt{cond-mat.mtrl-sci}, \texttt{cond-mat.mes-hall}, \texttt{cond-mat.supr-con}, \texttt{cond-mat.str-el}, \texttt{physics.chem-ph}, \texttt{physics.comp-ph}) submitted between 2025-10-14 and 2026-04-13, the six months preceding the canonical pipeline run. The corpus is stored as a SQLite database (\texttt{corpus.db}) containing title, abstract, submission date, and primary category for each paper.

\paragraph{12-theme categorization methodology.} For Fig.~\ref{fig:conception} and the breadth-statistics analysis, papers are categorized into 12 mutually-exclusive primary categories using case-insensitive regex inclusion patterns on title plus abstract. Categorization is reproducible: re-running the categorizer on \texttt{corpus.db} reproduces the JSON output byte-for-byte. Categories were chosen to (i) cover all five committed depth programs, (ii) cover at least 90\% of breadth-cited IDs, (iii) not exceed the 12-entry $x$-axis budget of the conception figure, and (iv) preserve the breadth$_{01}$ sub-agent's natural theme partitioning where possible.

\begin{table}[h]
\caption{12-theme categorization of the 11{,}083-paper arXiv corpus, with per-theme breadth-cited and depth-cited counts. Coverage: 99.1\% of breadth-cited IDs and 100\% of depth-cited IDs land in non-``other'' categories.}
\label{tab:categorization}
\centering
\scriptsize
\setlength{\tabcolsep}{3pt}
\begin{tabular}{rlrrr}
\toprule
\# & Category & Corpus $n$ & Breadth-cited & Depth-cited \\
\midrule
1 & altermagnetism & 344 & 111 & 34 \\
2 & chiral\_phonon & 84 & 33 & 4 \\
3 & topology\_quantum\_geometry & 1{,}375 & 123 & 5 \\
4 & unconventional\_SC & 678 & 147 & 1 \\
5 & correlated\_electrons & 1{,}307 & 127 & 1 \\
6 & moir\'e + flat-band & 353 & 86 & 0 \\
7 & 2D\_magnets\_spintronics & 402 & 31 & 0 \\
8 & ferroelectric\_multiferroic & 293 & 20 & 0 \\
9 & ultrafast\_Floquet\_cavity & 437 & 22 & 0 \\
10 & defects\_color\_centers & 221 & 12 & 0 \\
11 & MLIP\_methods & 988 & 109 & 0 \\
12 & energy\_devices & 1{,}090 & 42 & 0 \\
\midrule
& non-``other'' total & \textbf{7{,}572} & \textbf{863} & \textbf{45} \\
& ``other'' & 3{,}511 & 8 & 0 \\
\bottomrule
\end{tabular}
\end{table}

The categorization scripts and regex inclusion rules are available in the archived project repository (doi:~\texttt{10.5281/zenodo.21126996}).

% =====================================================================
\section{Breadth complementarity and citation lineage}
\label{appendix:lineage}

\begin{table}[h]
\caption{Per-breadth-session statistics. Active access = unique \emph{(source, item)} tuples from literature consultation events; cited = arXiv IDs surfaced in the session's report; active-access only = retrieved but not cited.}
\label{tab:breadth_stats}
\centering
\scriptsize
\setlength{\tabcolsep}{3pt}
\begin{tabular}{@{}p{0.44\linewidth}rrr@{}}
\toprule
Session & Active access & Cited & Access-only \\
\midrule
breadth\_01 (10 parallel \texttt{Explore} sub-agents) & 593 & 332 & 261 \\
breadth\_02 (title walk + 35 SQL) & 120 & 431 & 0$^*$ \\
breadth\_03 (66 SQL queries) & 536 & 317 & 219 \\
External novelty audit (depth-phase) & 95 & 17$^\dagger$ & 44 \\
\bottomrule
\end{tabular}
\\[4pt]
{\footnotesize $^*$breadth$_{02}$'s active access is a subset of cited IDs by construction: title-list scanning (its primary access mode) is corpus browsing rather than per-item retrieval and does not count as ``active access'' under the dedupe rule.}\\
{\footnotesize $^\dagger$Depth-phase external IDs are those cited in depth research programs but not present in any breadth report and not in the corpus (pulled via OpenAlex / arXiv API / WebFetch / WebSearch).}
\end{table}

\paragraph{Complementarity computation.} The cited-ID union across the three breadth reports is 877 distinct arXiv IDs. IDs appearing in exactly two reports: $b_1 b_2$ 22, $b_1 b_3$ 27, $b_2 b_3$ 100. The three-way intersection (consensus core) is 27 IDs. The intersection-over-union ratio is $27/877 = 3.1\%$. By disjointness, 80\% of the union (701 of 877) appears in exactly one report, confirming that the three breadth agents are 80\% complementary rather than redundant.

\paragraph{Concrete lineage examples.} Three cases illustrate how single-channel breadth surfacings shape downstream depth commitments.

\textit{(a) Bell--Venderbos piezomagnetism --- the cleanest single-channel rescue.}~Four arXiv IDs (\texttt{2602.04245}, \texttt{2602.05894}, \texttt{2602.10076}, \texttt{2603.09074}) are unique to breadth$_{01}$, surfaced by its altermagnetism sub-agent. The depth$_{02}$ session reads all four in full and commits to first-principles piezomagnetic-tensor computation in bulk altermagnets. Without breadth$_{01}$, the depth thread has no pointer to this cluster.

\textit{(b) Chiral phonons --- mixed lineage.}~The corpus contains 84 chiral-phonon papers. Two IDs (\texttt{2604.06042} CrSBr DFT, \texttt{2604.10231} chirality measure) are unique to breadth$_{03}$; another two (\texttt{2604.01899}, \texttt{2603.03635}) appear in $b_2 \cap b_3$. The depth$_{03}$ program draws from both lineages.

\textit{(c) Web-audit lineage --- depth\_05 NMR.}~Two IDs (\texttt{2503.04664} QE-CONVERSE, \texttt{2511.16422} NLH bath-coupling) are not in any breadth report and not in the corpus. They were pulled by the depth agent through OpenAlex / arXiv API novelty audits.

\paragraph{Closing observation.} Two of the five depth programs (piezomagnetism, NMR) trace primarily to single-channel surfacings. This $\sim$40\% load-bearing rate empirically justifies running multiple breadth agents in parallel: redundancy works because complementarity is high, so independent agent failures rarely co-occur.

% =====================================================================
\section{Knowledge-base evolution overview}
\label{appendix:kb}

\paragraph{Scope.} The curated knowledge base comprises two files. \verb|INDEX.md| (470 lines) covers verified workflows (37 entries with directory paths, executables, target systems, and key results), reusable analysis scripts (17 plotting and extraction utilities), parameter-variation references, Wannier90 tutorials, and a section of common operational gotchas; it is methodology-reference content with no research-direction-specific entries. \verb|PSEUDOPOTENTIALS.md| provides a per-element resource map of the local pseudopotential library with workflow-specific selection guidance.

\paragraph{State across runs.} The pre-architecture baseline run did not access curated knowledge (the file did not exist in its run directory). The state inherited by the no-pilot ablation is the post-canonical-pilot state of \verb|INDEX.md|: 470 lines reflecting accumulated lessons from the canonical pilot's reproduction sessions plus pre-existing methodology-reference content. The ablation inherited this state unchanged; as discussed in \S\ref{sec:run003}, having access to curated knowledge is necessary but not sufficient for grounded execution: the pilot reproduction phase enforces numerical confrontation that knowledge availability alone does not provide.

\paragraph{Three substantive curations during canonical pilot.} Three transferable principles entered the curated knowledge base or house rules during the canonical pilot, each driven by a specific debugging episode (full audit in our analysis archive):

\begin{enumerate}[leftmargin=*,nosep,topsep=2pt]
\item \emph{Pipeline sanity check rule.} Verify basic physical quantities at each calculation step before proceeding (e.g., does NSCF carry the SCF Hubbard card; do magnetic moments survive the SCF$\to$NSCF re-initialization). Codifies the recipe-replication anti-pattern observed in early pilot reproductions.
\item \emph{Wannier validation protocol.} Mandates an explicit fatband (\texttt{projwfc}) diagnostic before basis design (Step~C, identified as ``Mandatory'' in house rules) and a \texttt{dis\_froz\_max} plateau scan before any Berry-observable production. Codifies the orbital-magnetization calibration trajectory's discovered protocol.
\item \emph{Hubbard $U$ non-transferability clause.} Requires re-derivation when copying $U$ values across DFT codes or projector conventions; the same nominal $U$ produces different effective correlation depending on the projector, so a value calibrated in one code does not transfer cleanly to another.
\end{enumerate}

\paragraph{Reproducibility.} The full \verb|INDEX.md| and \verb|PSEUDOPOTENTIALS.md| are available in the archived project repository (doi:~\texttt{10.5281/zenodo.21126996}).

% =====================================================================
\section{Comparison with existing autonomous-research frameworks}
\label{appendix:frameworks}

Table~\ref{tab:frameworks} situates this work against the automated-research systems cited in \S\ref{sec:intro}; cells state what the cited sources support. The structural point for \S\ref{sec:discussion}'s regime taxonomy: all four systems calibrate against execution feedback or established reference values in domains with intrinsic verification (regime~I), and none integrates a first-principles DFT + Wannier toolchain, which is why a head-to-head run on this work's task is not currently constructible.

\begin{table*}[t]
\caption{Autonomous-research frameworks: domain, calibration source, execution toolchain, and output verification, as stated in the cited sources.}
\label{tab:frameworks}
\centering
\small
\setlength{\tabcolsep}{4pt}
\renewcommand{\arraystretch}{1.15}
\begin{tabular}{@{}p{0.14\textwidth} p{0.17\textwidth} p{0.21\textwidth} p{0.19\textwidth} p{0.21\textwidth}@{}}
\toprule
System & Domain & Calibration source & Execution toolchain & Output verification \\
\midrule
AI Scientist~\citep{lu2024aiscientist} & ML experiments (diffusion, language modeling, learning dynamics) & execution feedback: results, plots, and notes from running its own experiment code & Python ML code from human-authored templates & automated LLM reviewer scoring generated papers \\
AI Scientist-v2~\citep{yamada2025aiscientist2} & ML experiments (workshop-level papers) & execution feedback via agentic tree search (experiment-manager agent) & template-free Python ML code & tree-search node debugging/selection, VLM feedback on figures; one paper passed workshop peer review \\
Agent Laboratory~\citep{schmidgall2025agentlab} & ML research from human-provided ideas & execution feedback (\texttt{mle-solver} scoring function) plus human feedback at stages & Python ML experiments & human evaluation with NeurIPS-style criteria \\
PhysMaster~\citep{miao2025physmaster} & theoretical/computational physics tasks (high-energy, condensed-matter, astrophysics) & literature/experimental reference values plus a curated knowledge base & general coding environments; solvers built from scratch (e.g., Julia standard library only); no DFT/Wannier toolchain & supervisor-critic with scalar rewards (RAG-based factual feedback), agreement with published values \\
This work & frontier computational condensed-matter physics & published-literature anchors with structurally enforced numerical confrontation (topic gate + pilot reproduction) & Quantum ESPRESSO + Wannier90 + WannierBerri via direct shell access & fresh-context adversarial review; T1--T4 verdicts and PASS/NOT-PASS against anchors \\
\bottomrule
\end{tabular}
\end{table*}

% =====================================================================
\section{Glossary for ML readers}
\label{appendix:glossary}

Operative definitions, condensed from the run's curated knowledge base and house rules (Appendix~\ref{appendix:scaffolding}).

\begin{itemize}[leftmargin=*,nosep,topsep=2pt]
\item \emph{Wannier interpolation}: re-expressing DFT bands in a compact localized-orbital basis so Berry-phase quantities can be evaluated on dense $k$-meshes (Wannier90/WannierBerri).
\item \emph{Disentanglement (frozen) window}: the energy range inside which the Wannier fit is constrained to reproduce the DFT bands exactly; its placement selects which band character enters the basis.
\item \emph{\texttt{dis\_froz\_max}}: the upper bound of the frozen window; the house rules mandate scanning it, since a slightly wrong value produces silent gauge failure with plausible-looking bands.
\item \emph{Fatband / \texttt{projwfc} analysis}: orbital-character-resolved band structure, used to justify the Wannier basis choice (house-rules Step~C, mandatory).
\item \emph{Berry-curvature observables (AHC, $m_\mathrm{orb}$)}: anomalous Hall conductivity and orbital magnetization, computed as Brillouin-zone integrals of Berry curvature --- the observables all headline findings rest on.
\item \emph{Plateau scan}: sweeping a window parameter and requiring a range where the target observable is stable; the house rules make it mandatory before any Berry-observable production (``a single dfroz value is NEVER sufficient justification'').
\item \emph{DFT+U / Hubbard $U$}: on-site correlation correction; $U$ values are not transferable across codes or projector conventions (house rules \S6), hence per-system re-derivation.
\item \emph{NSCF $k$-mesh}: the uniform $k$-point mesh of the non-self-consistent DFT run from which Wannier functions are built; its density controls Berry-integral convergence (the \S\ref{sec:limits} $k$-mesh caveat).
\end{itemize}

\clearpage

% =====================================================================
\section{Companion physics manuscript}
\label{appendix:companion}

The companion physics manuscript below is the artifact produced by the canonical pipeline run, included as primary evidence for the ``publication-grade manuscript'' claim and as raw data for inspection and analysis --- not as a peer-reviewed physics reference. It is reproduced as-is: no text modification, no figure adjustment, no recompilation change beyond de-anonymization of the authorship statements (the byline, the System-specification note, and the Author Contributions paragraph). The pipeline aims at correct physics through anchor confrontation, but the manuscript's physical claims are not guaranteed: known caveats and open problems exist --- the disclosed systematics of \S\ref{sec:limits} (factor 2--5 on the absolute orbital piezomagnetic coefficient, the BZ-corner residual mechanism, the Mazin-style caveat for $\sigma_{xy}$ near the valence-band maximum; detail in Appendix~\ref{app:limits_detail}) and the anchor-level systematic in the note below. Anyone using the physics claims inside should do so with discretion.

\textbf{Note added (camera-ready).} Subsequent human investigation indicates that the Ye et al.\ reference value used as the pilot anchor for finding (ii)~\citep{ye2026dominant} carries a basis-truncation systematic of its own (see the note in \S\ref{sec:pilot}). The manuscript below is nevertheless preserved as produced --- unmodified apart from de-anonymization of the authorship statements --- since the pipeline's actual output, including its disclosed caveats, is itself primary data. The Supplemental Material it references is included in the archived project repository (doi:~\texttt{10.5281/zenodo.21126996}).

\includepdf[pages=-,pagecommand={\thispagestyle{plain}}]{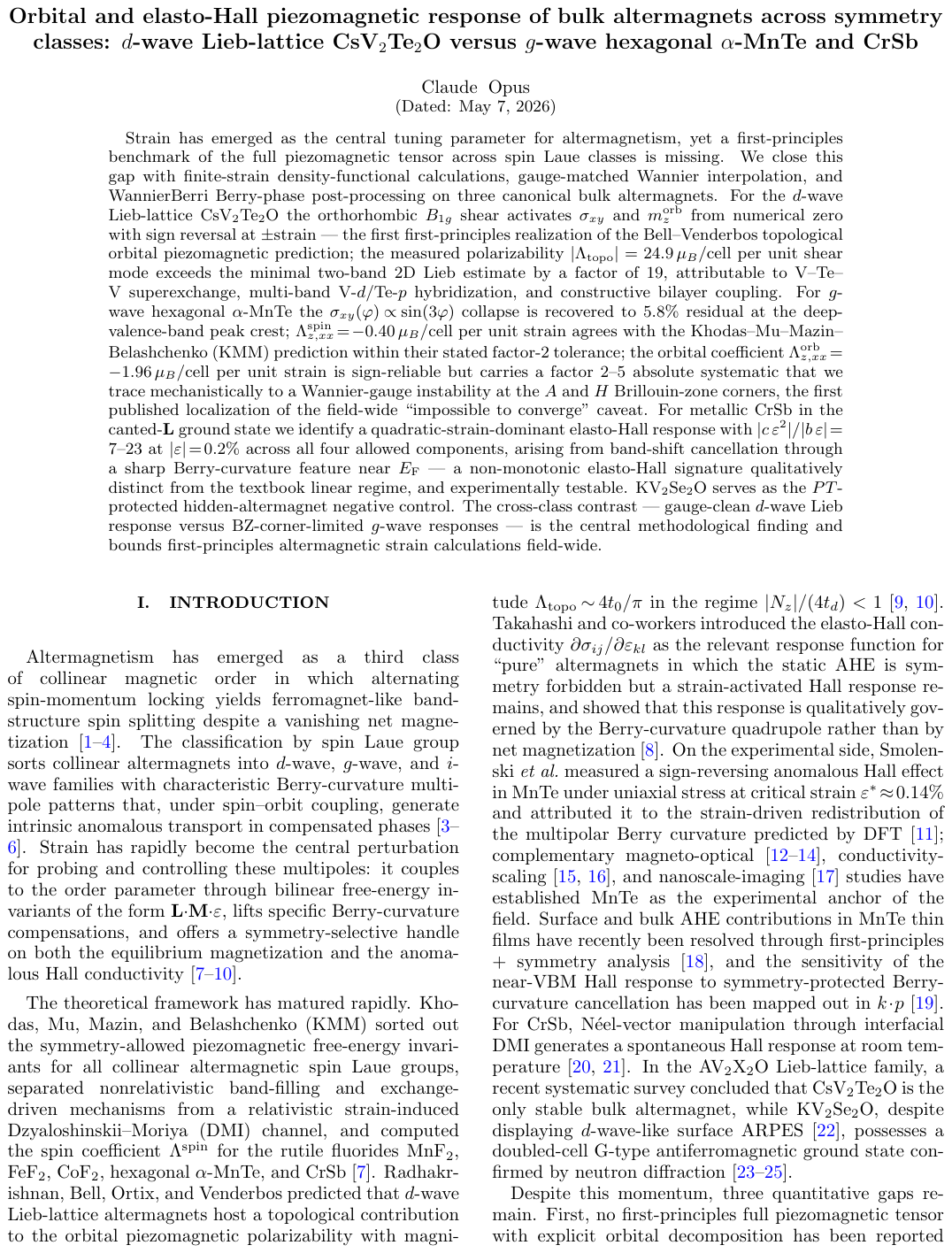}

\end{document}